\pdfoutput=1
\documentclass[sigconf]{acmart}
\usepackage{subfigure}

\usepackage{amssymb}
\usepackage{booktabs}
\usepackage{multirow}
\usepackage{graphicx}
\usepackage{amsmath}
\usepackage{color}
\usepackage{bbm}
\usepackage[normalem]{ulem}
\usepackage{microtype}
\usepackage[ ruled]{algorithm2e}
\usepackage{algorithmic}
\SetKwRepeat{Do}{do}{while}
\usepackage[ruled]{algorithm2e}
\usepackage{amsmath,caption}
\usepackage[hang,flushmargin]{footmisc}
\usepackage{lipsum}
\usepackage{setspace}
\usepackage{float}
\usepackage{stfloats}
\usepackage{multirow}
\usepackage{pifont}
\usepackage{balance}

\AtBeginDocument{%
  \providecommand\BibTeX{{%
    \normalfont B\kern-0.5em{\scshape i\kern-0.25em b}\kern-0.8em\TeX}}}

\newlength\savewidth\newcommand\shline{\noalign{\global\savewidth\arrayrulewidth
  \global\arrayrulewidth 1pt}\hline\noalign{\global\arrayrulewidth\savewidth}}

\definecolor{xinyu}{rgb}{0.5,0.8,0.7}
\definecolor{pink}{RGB}{219, 41, 145}

\makeatletter
\newcommand{\algorithmfootnote}[2][\footnotesize]{%
  \let\old@algocf@finish\@algocf@finish
  \def\@algocf@finish{\old@algocf@finish
    \leavevmode\rlap{\begin{minipage}{\linewidth}
    #1#2
    \end{minipage}}%
  }%
}

\renewcommand{\thefootnote}{\fnsymbol{footnote}}

\copyrightyear{2022}
\acmYear{2022}
\setcopyright{acmlicensed}\acmConference[MM '22]{Proceedings of the 30th ACM International Conference on Multimedia}{October 10--14, 2022}{Lisboa, Portugal}
\acmBooktitle{Proceedings of the 30th ACM International Conference on Multimedia (MM '22), October 10--14, 2022, Lisboa, Portugal}
\acmPrice{15.00}
\acmDOI{10.1145/3503161.3548108}
\acmISBN{978-1-4503-9203-7/22/10}
\begin{document}
\fancyhead{}

\title{IDEA: Increasing Text Diversity via Online Multi-Label \\ Recognition for Vision-Language Pre-training\vspace{-0.8em}}

\renewcommand{\shorttitle}{IDEA: Increasing Text Diversity via Online Multi-Label Recognition for Vision-Language Pre-training}

\author{Xinyu Huang$^{1}$, \ \ Youcai Zhang$^{2}$, \ \ Ying Cheng$^{3}$, \ \ Weiwei Tian$^{3}$, \ \ Ruiwei Zhao$^{3}$,\ \ Rui Feng$^{1,3,4*}$,  \ \ Yuejie Zhang$^{1*}$, \ \ Yaqian Li$^{2}$,\ \ Yandong Guo$^{2}$,\ \ Xiaobo Zhang$^{4,*}$}

\affiliation{\institution{$^{1}$School of Computer Science, Shanghai Key Laboratory of Intelligent Information Processing, Fudan University, China}\country{}}
\affiliation{
\institution{$^{2}$ OPPO Research Institute, China \quad\quad $^{3}$ Academy for Engineering and Technology, Fudan University, China\country{}}
}
\affiliation{
\institution{$^{4}$ Children’s Hospital of Fudan University, National Children’s Medical Center, China\country{}}
}

\affiliation{\institution{\{xinyuhuang20, chengy18, wwtian20, rwzhao, fengrui, yjzhang\}@fudan.edu.cn} \institution{\{zhangyoucai, liyaqian, guoyandong\}@oppo.com \quad zhangxiaobo0307@163.com}\country{}}

\renewcommand{\shortauthors}{Xinyu Huang et al.}

\newcommand\blfootnote[1]{%
\begingroup
\renewcommand\thefootnote{}\footnote{#1}%
\addtocounter{footnote}{-1}%
\endgroup
}
\vspace{3.0em}
\begin{abstract}
Vision-Language Pre-training~(VLP) with large-scale image-text pairs has demonstrated superior performance in various fields. However, the image-text pairs co-occurrent on the Internet typically lack explicit alignment information, which is suboptimal for VLP.
Existing methods proposed to adopt an off-the-shelf object detector to utilize additional image tag information. However, the object detector is time-consuming and can only identify the pre-defined object categories, limiting the model capacity. Inspired by the observation that the texts incorporate incomplete fine-grained image information, we introduce IDEA, which stands for increasing text diversity via online multi-label recognition for VLP. IDEA shows that multi-label learning with image tags extracted from the texts can be jointly optimized during VLP. Moreover, IDEA can identify valuable image tags online to provide more explicit textual supervision. Comprehensive experiments demonstrate that IDEA can significantly boost the performance on multiple downstream datasets with a small extra computational cost. Public code will be available at: \textit{\urlstyle{}{https://github.com/xinyu1205/IDEA-pytorch}}.


\vspace{-1.4em}

\blfootnote{ $^{*}$ indicates corresponding authors.} 

\end{abstract}



\maketitle




\begin{small}
\setlength{\parskip}{-1.0em} 
\setlength{\parindent}{0pt}
\begin{spacing}{1}
\textbf{ACM Reference Format:} \\
Xinyu Huang, Youcai Zhang, Ying Cheng, Weiwei Tian, Ruiwei Zhao, Rui Feng, Yuejie Zhang, Yaqian Li, Yandong Guo, Xiaobo Zhang. 2022. IDEA: Increasing Text Diversity via Online Multi-Label Recognition for Vision-Language Pre-training. In \emph{Proceedings of the 30th ACM International Conference on Multimedia (MM '22), October 10-14, 2022, Portugal, Lisboa.} ACM, New York, NY, USA, 10 pages. https://doi.org/10.1145/3503161.3548108
\end{spacing}
\end{small}

\begin{figure} [t]
\centering
  \includegraphics[width=1.0\linewidth]{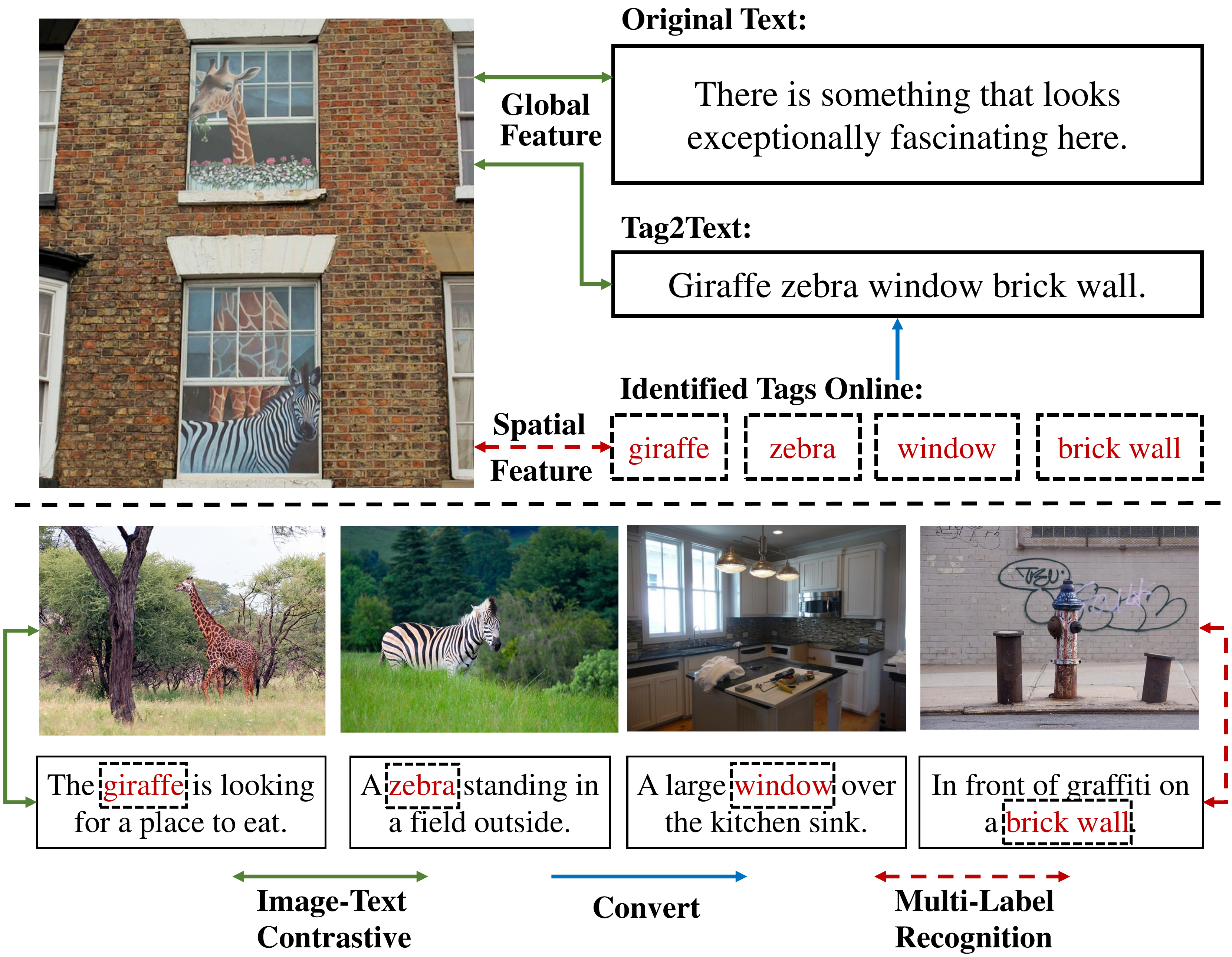}
  \caption{ 
Training vision-language models with image-text pairs co-occurrent on the Internet is suboptimal, as such supervision typically lacks explicit alignment information~(the original text in upper row). We propose IDEA to provide more explicit supervision~(including multiple valuable tags and texts composed by multiple tags). Our IDEA jointly trains multi-label recognition with tags from texts~(shown in lower row) and identifies additional tags online~(the identified tags online in upper row).
}
\vspace{-2.0mm}
  \label{fig:intro}
\end{figure}

\section{Introduction}



The two-stage training paradigm of "pre-training \& fine-tuning" has gradually become a standard scheme in deep learning. Pre-training on large-scale datasets can significantly improve the performance and generalization of the model. Actually, large-scale data not only helps define an approximation to the target problem, but also is a necessary condition to ensure asymptotic convergence~\cite{li2022vision}.
In the field of Vision-Language Pre-training~(VLP), CLIP~\cite{radford2021learning} and ALIGN~\cite{jia2021scaling} collected millions of image-text pairs for learning visual representation from natural language supervision, which has been proven to transfer to various downstream tasks, such as vision-and-language tasks~\cite{shen2021much}, image tasks~\cite{radford2021learning,gu2021zero,rao2021denseclip}, and video tasks~\cite{ju2021prompting,luo2021clip4clip,fang2021clip2video}, etc. They directly align visual-linguistic features by the Image-Text Contrastive (ITC) loss and can scale up large-scale datasets benefiting from the high training efficiency.

Although image-text pairs can be collected in large quantities from the Internet, the weakly textual supervision is suboptimal for VLP. The subjective texts generally cannot describe the objective image information comprehensively, which can only provide weakly supervised information. The example in Figure~\ref{fig:intro}~(upper row) illustrates that the original text simply describes some abstract or implicit concepts (\textit{e.g., something, fascinating}) of the image while lacking explicit alignment information  (e.g., \textit{giraffe, zebra, window, etc.}). We argue that the lack of textual supervision has been largely overlooked, while the attention is mainly drawn to the incredible performance by continuously scaling up the amount of data. 



Recently, BLIP~\cite{li2022blip} designed a dataset-bootstrapping method called CapFilt to produce more synthetic captions by pre-training the generation-based task with image-text pairs. BLIP found that more diverse captions yielded larger gains for VLP models.
Besides, existing methods~\cite{li2020oscar,zhang2021vinvl} proposed that taking advantage of the visual spatial feature corresponding to the tag information could effectively improve the ability of VLP models. However, both of them rely on the off-the-shelf object detectors~(e.g.,\textit{ Faster RCNN~\cite{ren2015faster}}) to extract tag information on the visual side. 
This can result in the following problems: (i) The ability of the object detectors has an upper limit, that is, identifying all of the pre-defined object categories;
(ii) The imperfect object detectors generally remain frozen and cannot be optimized during VLP, which limits the model capacity; and (iii) The object detectors can be time-consuming, while training efficiency is crucial to scaling natural language supervision~\cite{radford2021learning}.


Inspired by the observation that the texts of the image-text pairs contain incomplete fine-grained image information~(e.g., \textit{tags, phrases, etc.}), we exploit this information to enrich supervised signals with textual information for improving visual representation learning. Concretely, we design a novel and effective VLP framework named \textbf{IDEA}, which stands for \textit{increasing text diversity via online multi-label recognition for vision-language pre-training}.
IDEA applies Image-Text Contrastive learning~(ITC) with the global features obtained by image and text encoders to align the visual-linguistic features.  
Meanwhile, IDEA utilizes the visual spatial feature to accomplish Multi-label Recognition~(MLR), which aims to identify multiple tags contained in an image. 
Compared with previous methods, IDEA mainly has the following advantages:
\begin{itemize}
\item IDEA directly utilizes image tags~(\textit{including objects, scenes, attributes, actions, etc.}) extracted from the texts for MLR training, so that no manual annotation is required and MLR can be jointly trained during VLP. MLR only requires an additional recognition head with a small extra computational cost without reducing the training efficiency.

\item The combination of ITC with MLR can boost performance mutually, indicating that although they are two diverse vision tasks, both can essentially optimize the representation of the visual backbone.

\item Moreover, IDEA can identify valuable image tags online to provide more explicit alignment information. As shown in Figure~\ref{fig:intro}, the identified tags online~(e.g., \textit{"giraffe", "zebra"}) can be converted into a new semantically rich textual description.

\end{itemize}

Comprehensive experiments demonstrate the effectiveness of IDEA. Specifically, IDEA significantly improves the performance on 7 out of 9 visual datasets when transferred to downstream tasks. With 213K and 3M data pre-trained, IDEA has 2.18\% and 1.02\% improvement on ImagenNet-1k zero-shot top1 accuracy, respectively. Furthermore, the analysis experiment illustrates that the identified online image tags with high accuracy can provide practical and effective text supervision. IDEA shows that taking advantage of the fine-grained image information implied in the texts can further boost the latent capacity of the model.

\section{Related Work}

\subsection{Vision-Language Pre-training}
The development of Vision-Language Pre-training~(VLP) has seen a significant boost in the domain of vision-language intelligence, which aims to learn versatile visual-linguistic features. According to the modality interaction scheme, VLP can be divided into two categories~\cite{du2022survey}. (i)~The fusion encoder models generally adopt a deep fusion encoder with co-attention~\cite{lu2019vilbert,tan2019lxmert,li2021align,dou2021empirical,li2021align} or merged attention~\cite{li2019visualbert,chen2020uniter,li2020oscar,zhang2021vinvl,huang2021seeing} for modality interaction. The pre-training tasks of these models typically include Image-Text Matching~(ITM), Masked Language Modeling~(MLM), etc. (ii)~The dual encoder models directly calculate cosine similarity between the modality features. Then the InfoNCE loss~\cite{van2018representation} is used to align matched image-text pairs as positive pairs, while other unmatched image-text pairs as negative pairs are pulled away. This pre-training task is called Image-Text Contrastive learning~(ITC), whose representative works include CLIP~\cite{radford2021learning}, ALIGN~\cite{jia2021scaling}, etc. The dual encoder models can scale up visual-linguistic representation learning with high training efficiency. The visual models supervised by natural language can be transferable to various downstream tasks, such as vision-language tasks~\cite{shen2021much,radford2021learning}, image tasks~\cite{radford2021learning,gu2021zero,rao2021denseclip} and video tasks~\cite{ju2021prompting,luo2021clip4clip,fang2021clip2video}, etc. In this paper, our work focuses on boosting the performance of the second category without reducing the training efficiency.

Since the co-occurrent texts generally cannot fully describe the images, they can only provide weakly supervised information. To address this issue, BLIP~\cite{li2022blip} proposed a dataset bootstrapping method called CapFilt to produce more synthetic captions and found more diverse captions yield larger gains for VLP models. In particular, the Captioner module in CapFilt required pre-training the generation-based task with image-text pairs first. Based on the discovery that Bag-of-Words~(BoW) captions can also provide fabulous supervisory information~\cite{tejankar2021fistful}, we consider enriching text diversity via valuable image tag information. OSCAR~\cite{li2020oscar} and VinVL~\cite{zhang2021vinvl} utilized image tag information and demonstrated outstanding results in various vision-language tasks. However, these additional tags are obtained from the off-the-shelf object detectors, which can be time-consuming and only identify pre-defined object tags. Moreover, the object detectors generally remain frozen and cannot be optimized during VLP, limiting the model capacity. In this paper, we consider obtaining valuable image tag information~(\textit{including objects, scenes, attributes, actions, etc.}) via Multi-Label Recognition~(MLR). To the best of our knowledge, our work is the first attempt that introduces MLR into VLP with tag information implied in the texts.

\begin{figure*} [t]
\centering
  \includegraphics[width=0.98\linewidth]{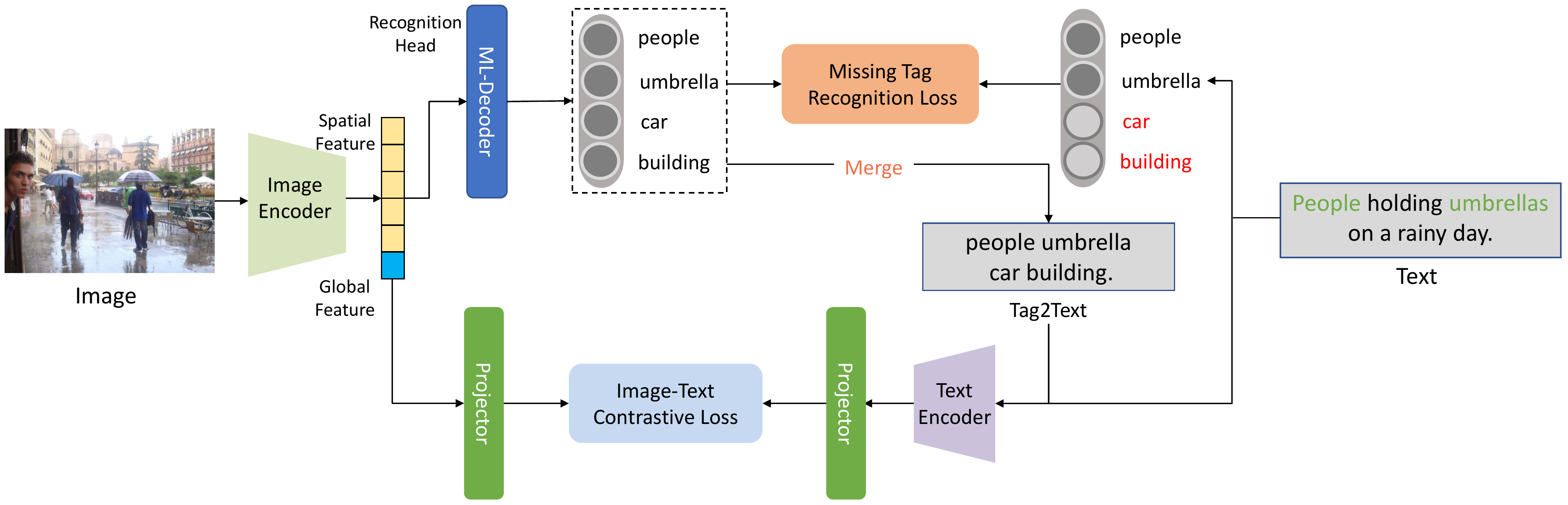}
  \caption{An overview of our proposed IDEA framework. IDEA applies Image-Text Contrastive learning~(ITC) with global features obtained by image encoder and text encoder, respectively. In addition, IDEA adopts ML-Decoder based on the Transformer decoder as the recognition head to complete Multi-Label Recognition~(MLR) with visual spatial feature. The tags are extracted directly from the texts with no manual annotation. The combination of ITC with MLR can improve performance mutually. Moreover, by leveraging missing tag recognition loss, IDEA can identify the missing tags~(e.g., \textit{"car", "building"}) online to provide more textual supervised signals for ITC.}
  \label{fig:model_architercture}
\end{figure*}

\vspace{-0.5em}
\subsection{Multi-Label Recognition}
Multi-Label Recognition~(MLR) aims to identify multiple tags for a given image. On the one hand, since these tags tend to appear in different image positions, Query2Label~\cite{liu2021query2label} adopted the transformer decoder~\cite{vaswani2017attention} as the recognition head to take full advantage of visual spatial feature. On this basis, ML-Decoder~\cite{ridnik2021ml} developed the transformer decoder, which alleviated the problem of high computational complexity when the number of categories was large. On the other hand, owing to the difficulty of annotating all the ground-truth tags for an image, many attempts have been made to explore MLR learning with missing tags. Durand \emph{et al.}~\cite{durand2019learning} proposed a multi-stage training scheme to correct missing tags. Cole \emph{et al.}~\cite{cole2021multi} designed the network structure ROLE to correct missing tags by training jointly the image classifier and the label estimator. Zhang \emph{et al.}~\cite{zhang2021simple} presented that the missing tags could be distinguished in the early training stage and proposed online Self-Spaced Loss Correction~(SPLC) to improve model performance without increasing the procedure and complexity.

Most existing multi-label datasets are human-annotated (e.g., \textit{COCO~\cite{chen2015microsoft}, VOC~\cite{everingham2011pascal}}), which can be expensive for construction. Considering the large quantities of image-text pairs available on the Internet, we explore MLR learning by directly extracting image tag information from the corresponding text.





 



\section{Approach}
\subsection{Framework}
Figure~\ref{fig:model_architercture} illustrates our IDEA pre-training paradigm. IDEA consists of an image encoder, a text encoder and a recognition head. 
With a standard ViT~\cite{dosovitskiy2020image} as the image encoder, an image $I$ is encoded into a sequence of embeddings $\{v_{cls},v_1,...,v_N\}$, where $v_{cls}$ is the class token indicating the  visual global embedding while the rest are visual spatial embeddings. Similarly, the corresponding text $T$ is encoded into the text encoder based on the Transformer structure to get the sequence embeddings $\{w_{cls},w_1,...,w_N\}$. 
The visual global embedding is used to align visual-linguistic features with the text, and the visual spatial embeddings are used to recognize tags located in different areas of the image. We adopt the transformer decoder based on self-attention as the recognition head to fully use visual spatial embeddings. By leveraging the missing tag recognition loss, IDEA can identify missing tags online to provide additional textual information. The pseudo-code of IDEA training scheme is provided in Appendix~\ref{sec:pseudo}.



\subsection{Multi-Label Recognition}
\noindent
\textbf{Extracting Tags from Texts.} We explore Multi-Label Recognition~(MLR) by leveraging the large quantities of available image-text pairs. Concretely, the image tags are extracted directly from the corresponding text according to the tag list without manual annotation. The tag list can be expanded arbitrarily. In this paper, we construct it based on the multi-label recognition dataset of OpenImages~\cite{kuznetsova2020open}, which contains 9,600 classes and covers a wide range of real-world concepts. We select the categories with higher frequency in texts to our tag list. More details about tag list construction can be found in Sec~\ref{tag_construction}. 


For the obtained tag list containing $C$ categories, there are noun tags~(e.g., \textit{"dog"}) and compound noun tags~(e.g., \textit{"hot dog"}). For the former, we perform tokenization and lemmatization of sentences using the WordTokenize and WordNetLemmatizer from the NLTK library~\cite{loper2002nltk}, and then match the individual words with the noun tags. For the latter, we only select the compound nouns without hypernyms and match them with sentences directly. If the compound noun tag includes the noun tag, we only select the compound noun tag.



\begin{figure}[t]
  \centering
  \includegraphics[width=.45\textwidth]{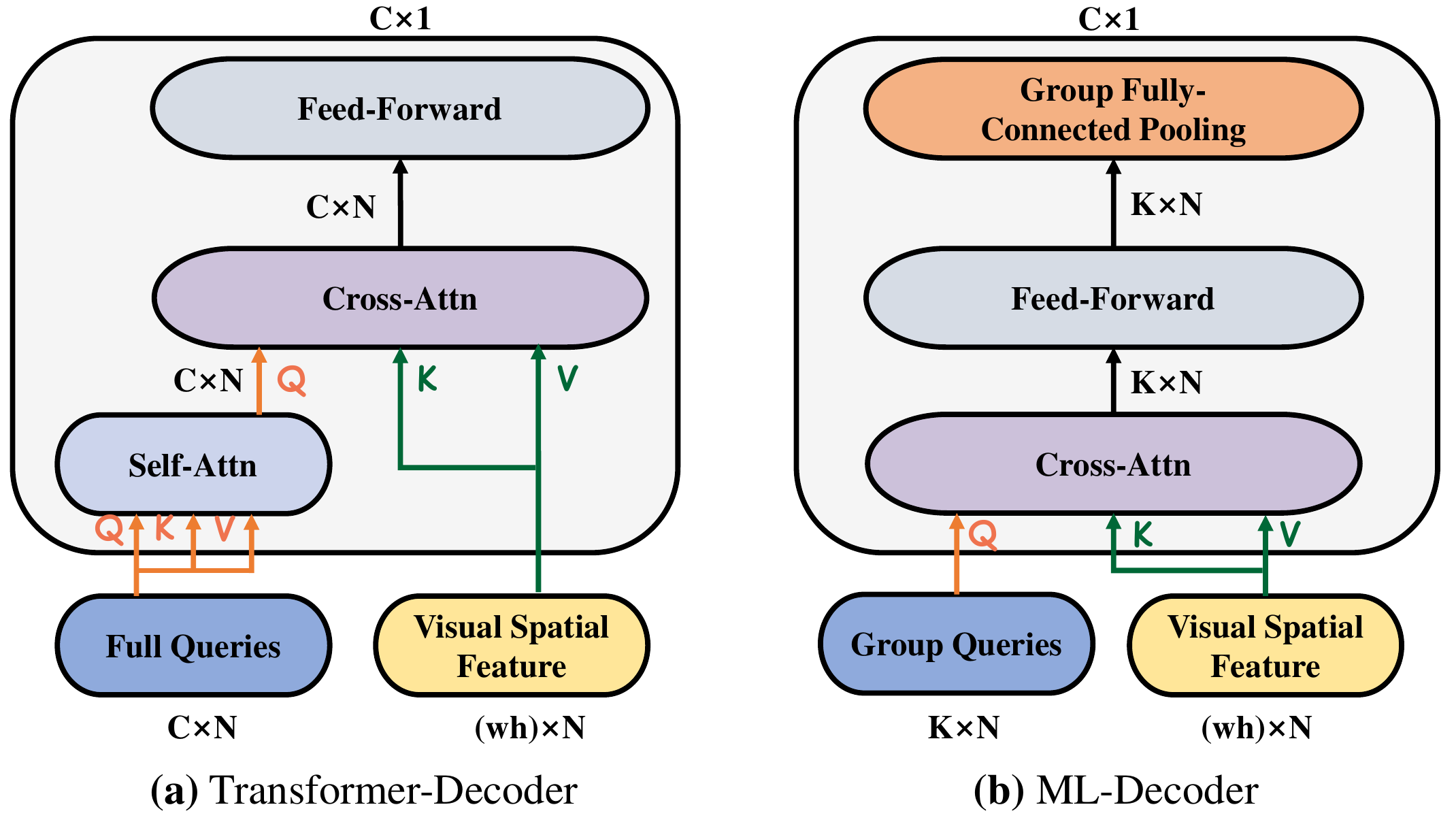} 
  \caption{Comparison of different multi-label recognition heads utilizing visual spatial feature. }
  \label{fig:mldecoder}
\end{figure}

\vspace{5pt}
\noindent
\textbf{Recognition with Spatial Feature.} The traditional single-label classification directly adopts the visual global feature $v_{cls}$ passing through the fully connected layer for classification. Unlike single-label classification, MLR aims to identify multiple labels with different contents in an image. Therefore, we apply ML-Decoder~\cite{ridnik2021ml} based on the Transformer decoder~\cite{liu2021query2label,vaswani2017attention} as the recognition head to take advantage of visual spatial feature.
As illustrated in Figure~\ref{fig:mldecoder}, the visual spatial embeddings $\{v_1,v_2,...,v_N\}$ are put into the ML-Decoder as keys \& values utilizing cross-attention with $N$ group queries $\{q_1,q_2,...,q_N\}$. The $N$ group queries can be random fixed values in that the learnable projection can transform them into any values~\cite{ridnik2021ml}. Then the outputs are passed through feed-forward layers and group fully connected pooling, and finally converted to logits $\{x_1,x_2,...,x_C\}$ corresponding to $C$ tags, where $x_i$ denotes the logit for \textit{i}-th category. Benefiting from the advancement on transformer decoder, the FLOPs of ML-Decoder recognition head are almost the same as Global Average Pooling, even if the categories number $C$ is huge.

\vspace{5pt}
\noindent
\textbf{Missing Tag Recognition Loss.} Since MLR demands to select multiple target categories from $C$ candidate categories, it is generally decomposed into multiple binary classification. Specifically, the probability for each category $\{p_1,p_2,...,p_C\}$ is obtained by normalizing the logits respectively with the sigmoid function as $p_{i} = \sigma(x_i) =  \frac{1}{1+e^{-x_i}}$. Then the binary classification loss is generally given by:
\begin{equation}
    \mathcal{L}_{mlr} = - \sum_{i=1}^C \left( y_i L^+_i + (1-y_i)L^-_i \right),
\end{equation}
where $y_i$ denotes the ground-truth of the \textit{i}-th class, $L^+_i$ denotes the positive sample loss, and $L^-_i$ denotes the negative sample loss. Due to the extremely long-tailed distribution of categories, we apply simple re-weighting to improve the performance of categories with few samples:
\begin{equation}
    \mathcal{L}_{mlr} = - \sum_{i=1}^C w_i\left( y_i L^+_i + (1-y_i)L^-_i \right),
\end{equation}
where $w_i$ is inversely proportional to the square root of category frequency. For simplicity, we neglect the subscript $i$ in $L^+_i$ and $L^-_i$ to $L^+$ and $L^-$ in the subsequent formula. The most commonly used Binary Cross Entropy~(BCE) loss in multi-label recognition is defined as:
\begin{equation}
\begin{cases}
    &L_{BCE}^+ = \log(p)    \\
    &L_{BCE}^- = \log(1-p)
\end{cases},
\end{equation}

The image tags extracted from the co-occurrent texts are generally incomplete, leading to the presence of the missing tags. For instance, \textit{"car"} and \textit{"building"} in Figure~\ref{fig:model_architercture}, which exist in the image yet ground-truth $y_i=0$, will be treated as negative samples to optimize in the wrong direction. 

Due to the symmetric design of BCE, positive and negative samples are treated equally. There exists severe imbalance between positives and negatives, especially when the number of categories $C$ is huge. In this case, most categories are negative for a given image. With the domination of negatives over positives, the model would preferentially optimize the negative sample loss. If a negative sample can still have a higher predicted probability in such situation, it has high precision of a missing tag. 
This chacteristic enables the predicted probability to identify missing tags even in the early stage of training~(e.g., \textit{after the $1st$ training epoch}), which is called Self-Paced Loss Correction~(SPLC): 
\begin{equation}
\begin{aligned}
\begin{cases}
    &L^+_{splc} = \log(p)    \\
    &L^-_{splc} = \mathbb{I}(p\leq \tau)\log(1-p) + (1-\mathbb{I}(p\leq \tau))\log(p)
\end{cases},
\end{aligned}
\end{equation}
where $\mathbb{I}(\cdot)\in\{0,1\}$ is the indicator function, and $\tau\in(0,1)$ is the threshold. When the predicted probability of a given negative sample $p>\tau $, a pseudo positive label would be generated to correct the direction of optimization.

\subsection{Cross-Modal Learning}
\noindent
\textbf{Converting Tags to Texts.} The missing tags identified by SPLC~(e.g., \textit{"car", "building" }in Figure~\ref{fig:model_architercture}) cannot only help optimize multi-label recognition better, but also can be directly converted into a new text to increase the text diversity corresponding to the image. We merge the recognized missing tags and original tags $[tag_1],[tag_2],...[tag_i]$ into a new text $T_{tag}$: [$tag_1$  $tag_2$ ... $tag_i$] corresponding to the image $I$. Then both $T$ and $T_{tag}$ can provide textual supervision for image $I$. If there are no additionally identified missing tags, we set $T_{tag}$ as an empty string.



\vspace{5pt}
\noindent
\textbf{Image-Text Contrastive Learning.}
We adopt Image-Text Contrastive learning~(ITC) to align the image and text features. For the global feature $v_{cls}$ of image $I$ and $w_{cls}$ of text $T'$ ($T'={T \circ T_{tag}}$ and $\circ$ denotes the concatenation function), the projector heads $g_v$ and $g_w$ are used to map them to the same dimension~(e.g., \textit{256-d}), respectively. 

Then the normalized features are used to calculate image-to-text similarity $s_{i2t}=g_v{(v_{cls})}^Tg_w(w_{cls})$ and text-to-image similarity $s_{t2i}=g_w{(w_{cls})}^Tg_v(v_{cls})$, respectively. The softmax-normalized similarities are given by:
\begin{equation}
p_{i2t} = \frac{exp(s_{i2t}/\pi )}{\sum_{m=1}^{M}exp(s_{i2t_m}/\pi)}
,
p_{t2i} = \frac{exp(s_{t2i}/\pi )}{\sum_{m=1}^{M}exp(s_{t2i_m}/\pi)}
\end{equation}
where $M$ is the batch size, and $\pi$ is a learnable temperature parameter. Since many images have additionally recognized texts $T_{tag}$, the similarity score cannot be regarded as an \textit{n vs.} 1 classification problem. We adopt the KL divergence loss to align the similarity matrix with the target values $y_{i2t}$ and $y_{t2i}$. Note that we set the target to 1 when the corresponding $T_{tag}$ is not an empty string.
The image-text contrastive loss is defined as: 
\begin{equation}
\mathcal{L}_{itc}=\frac{1}{2}[KL(p_{i2t},y_{i2t})+KL(p_{t2i},y_{t2i})]
\end{equation}

The full pre-training objective of IDEA is: 
\begin{equation}
\mathcal{L}=\mathcal{L}_{mlr} + \mathcal{L}_{itc}
\end{equation}

\section{Experiment}

\subsection{Pre-training Setup} 
\noindent
\textbf{Pre-training Datasets.} Following~\cite{chen2020uniter,li2019visualbert,tan2019lxmert,huang2021seeing}, we construct our pre-training data on two widely-used datasets, that is, COCO~\cite{chen2015microsoft} and Visual Genome\cite{krishna2017visual}~(VG). Since there are multiple texts to describe each image on average, we randomly discard these texts to verify the superiority of IDEA adequately. Finally, each image of COCO\&VG has one textual description,that is, COCO with ($\sim$ 113K) image-text pairs and VG with ($\sim$ 100K) image-text pairs. We also conduct experiments on an additional web dataset Conceptual Captions~\cite{sharma2018conceptual} with ($\sim$ 2.9M) image-text pairs~(CC-3M). Unless illustrating pre-trained with CC-3M, the default experimental results are pre-trained with COCO\&VG. 

\begin{figure}[t]
  \centering
  \includegraphics[width=.45\textwidth]{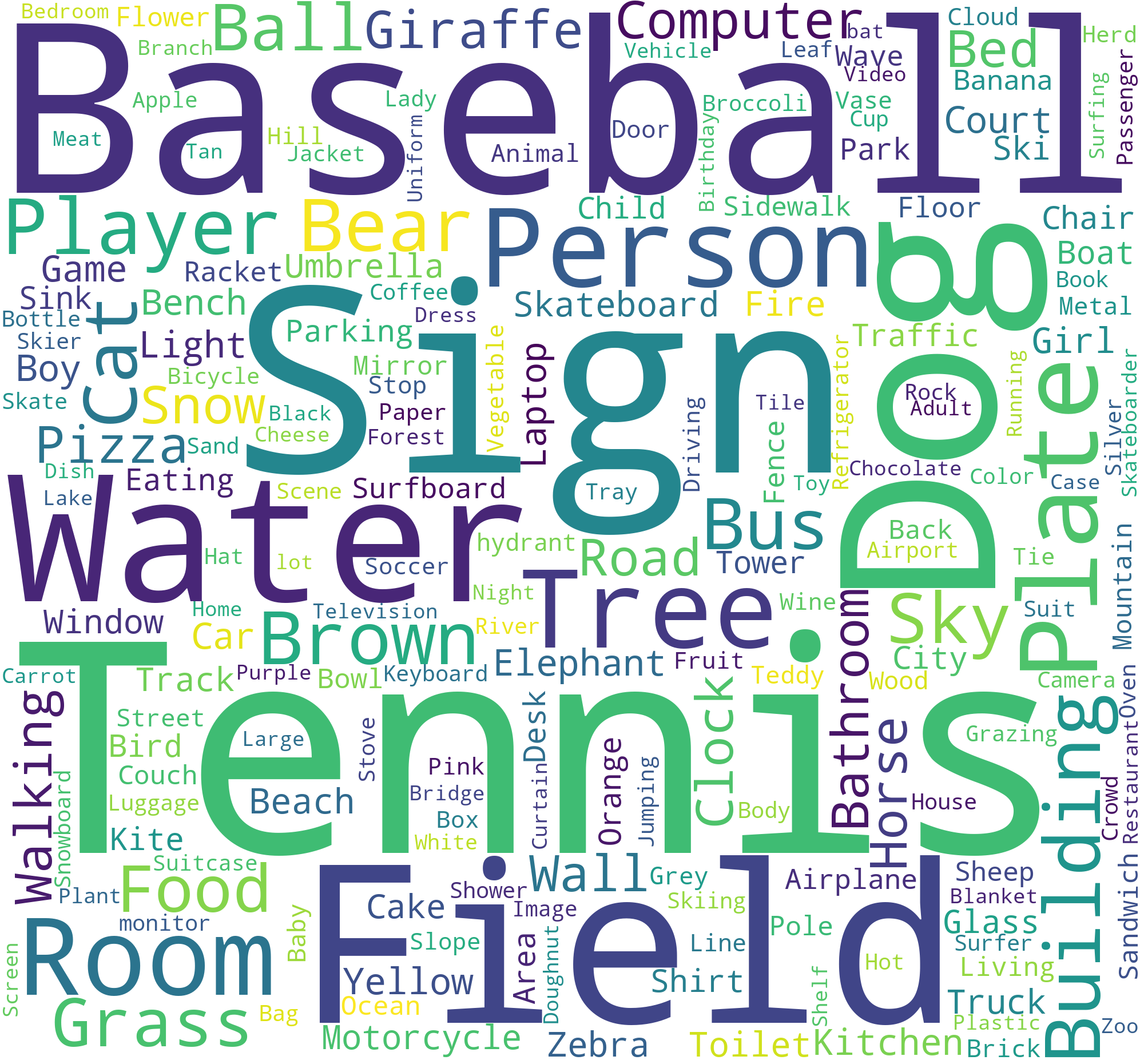} 
  \vspace{-0.5em}
  \caption{Illustration of the most-frequent categories in the tag list. The word size is proportional to the category frequency in the training set.}
  \vspace{-1.0em}
  \label{fig:word_cloud}
\end{figure}

\vspace{5pt}
\noindent
\textbf{Tag List Construction.} We build the tag list based on OpenImages~(V6)~\cite{kuznetsova2020open} to extract image tags from the co-occurrent texts. OpenImages is a large-scale visual database that includes 9,600 trainable image-level classes~(consisting of objects, scenes, attributes, actions, etc.) and 600 boxable classes~(only consisting of objects). We use the complete list of 9,600 classes to extract tags from captions and select the category tags with more than 5 samples. We also remove top most frequent tags since they can degrade the model performance even if they are correctly identified~(see Section~\ref{label2text_ablation_study}). The final tag list consists of 1,000/2,500 classes for COCO\&VG/CC-3M. An overview of the most frequent classes in COCO\&VG is shown in Figure~\ref{fig:word_cloud}.

Noting that some tags have polysemy, such as \textit{"coach"}, while others have synonyms, such as \textit{"ocean"} and \textit{"sea"}. We consider further cleaning of the tag list can promote the performance. Owing to this requiring more manual overhead, we leave it as future work.

\label{tag_construction}

\vspace{5pt}
\noindent
\textbf{Implementation Details.} We adopt ViT-B/16~\cite{dosovitskiy2020image} pre-trained on ImageNet~\cite{deng2009imagenet} with 85.5M parameters as the image encoder and BERT$_{base}$~\cite{devlin2018bert} with 123.7M parameters as the text encoder. We pre-train the model for 30 epochs by default with the batch size of 1,024 on 8 NVIDIA V100 GPUs. The input images are uniformly resized to $256\times256$, and the maximum length of the text is padded to 40. The optimizer is the AdamW with a weight decay of 0.02, and cosine schedule~\cite{loshchilov2016sgdr} is used with the max learning rate of 1e-4. Following \cite{li2021align}, we also apply RandAugment~\cite{cubuk2020randaugment} but remove color changes, since text and tags usually contain color information. The threshold $\tau$ is set to 0.6, and the changing epoch is set to 1st in SPLC.

\begin{figure*} [ht]
\centering
  \includegraphics[width=0.98\linewidth]{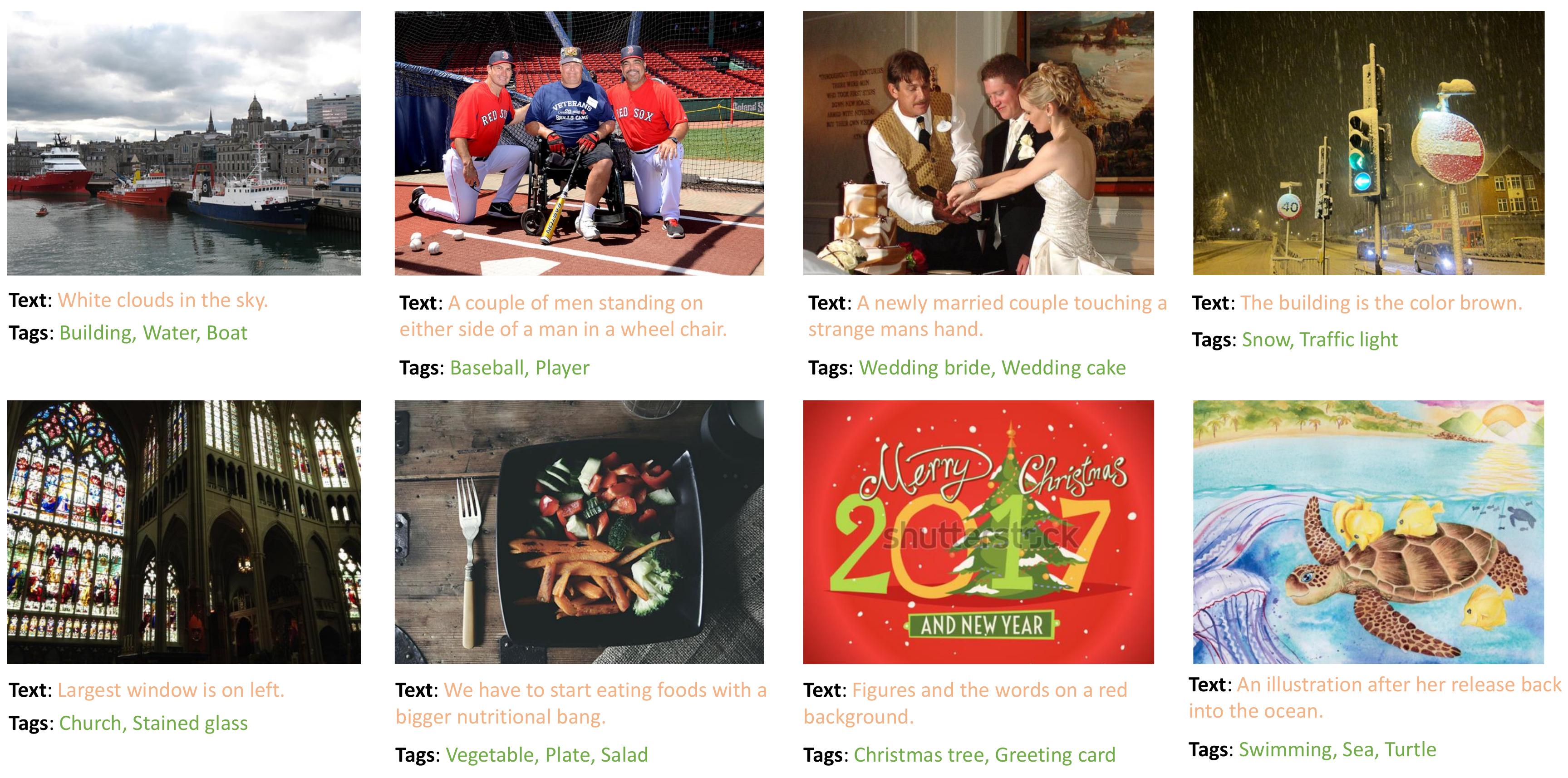}
  \vspace{-0.5em}
  \caption{Example results on COCO\&VG~(1st row) and CC-3M~(2st row). Text refers to the original co-occurrent texts with the image. Tags refer to the identified tags by IDEA, including objects, scenes, attributes, actions, etc. These image tags are entirely learned from the texts and recognized online.}
  \label{fig:visualization}
\end{figure*}

\subsection{Evaluation Setup}
\noindent
\textbf{Downstream Datasets.} Following~\cite{radford2021learning,tejankar2021fistful,li2021supervision}, we validate the effectiveness of our proposed method via zero-shot image classification evaluation on 9 different downstream datasets, including the most commonly used classification dataset ImageNet-1k~\cite{deng2009imagenet} and 8 fine-grained classification datasets, that is, CIFAR10~\cite{krizhevsky2009learning}, CIFAR100~\cite{krizhevsky2009learning}, Caltech101~\cite{fei2004learning},  Flowers102~\cite{nilsback2008automated}, Food101~\cite{bossard2014food}, DTD~\cite{cimpoi2014describing}, SUN397~\cite{xiao2010sun}, and OxfordPets~\cite{parkhi2012cats}.

\vspace{5pt}
\noindent
\textbf{Prompt Engineering and Ensembling.} To accomplish zero-shot evaluation, we calculate similarity by directly feeding the labels into text encoder, and select the label with the highest similarity. Following CLIP~\cite{radford2021learning}, we adopt prompt engineering and ensembling to further improve the performance. Specifically, the labels are put into different prompt templates such as {\rm{"a bad photo of a \{label\}"}} and {\rm{"a photo of many \{label\}"}}, and then averaged to assemble in the textual representation space. The top-1 accuracy and top-5 accuracy are used as the metrics to evaluate zero-shot classification performance.

\begin{table*}[ht]
\centering
\caption{Evaluation of the proposed methods on different downstream datasets. IN1k Zero-Shot refers to the zero-shot accuracy on ImageNet-1k. Average Zero-Shot refers to the average zero-shot accuracy on 9 datasets. Multi-label performance refers to the performance of multi-label recognition on OpenImages using the recognition head. The default experimental results are pre-trained on COCO\&VG and the results with (3M) are pre-trained on CC-3M. }
\label{tab:proposed_methods}
\begin{tabular}{c|c|c|c|cc|cc|ccc}
 &
   \multicolumn{1}{c|}{} &
  \multicolumn{1}{c|}{} &
  \multicolumn{1}{c|}{} &
  \multicolumn{2}{c|}{IN1k Zero-Shot} &
  \multicolumn{2}{c|}{Average Zero-Shot} &
  \multicolumn{3}{c}{Multi-Label Performance} \\
  Method & ITC & MLR & Tag2Text  & Top-1 & Top-5 & Top-1 & Top-5 & mAP & CP & CR \\  \shline
CLIP & \ding{51} &                &                  &14.54     & 31.18     &  18.95  & 35.61   &  -  &  -  &  -   \\ 
 - &    &    \ding{51}   &                   &  -    &  -    &  -  & -   &30.89    &63.11    & 10.67    \\ \hline
 \textbf{IDEA}* &  \ding{51} &    \ding{51}  &      &  15.30   & 31.43   &  18.91   &36.60    &\textbf{34.90}     & \textbf{65.48}    &  19.60   \\ 
 \textbf{IDEA} &  \ding{51} &    \ding{51}    &   \ding{51}    & \textbf{16.72}   & \textbf{34.04}  & \textbf{19.57}  & \textbf{37.70}  &  34.40 &63.27     & \textbf{21.32}    \\ \hline
CLIP~(3M)&\ding{51}    &       &       &    35.30         &   57.87  &34.13   &54.21    & -    &- &-  \\
\textbf{IDEA~(3M)}&\ding{51}     &    \ding{51}   &   \ding{51}  &  \textbf{36.32}    &  \textbf{58.66}   & \textbf{34.76}   &\textbf{55.02}    & 33.10    & 66.56    &14.09
\end{tabular}
\end{table*}

\vspace{5pt}
\noindent
\textbf{Multi-Label Recognition.} We also validate the Multi-Label Recognition~(MLR) performance of our model using the official OpenImages~\cite{kuznetsova2020open} test set. Since only part of the labels are selected during pre-training, we simply test on the data containing these labels. Following~\cite{zhang2021simple,ridnik2021ml,liu2021query2label}, mean Average Precision~(mAP), Overall F1-measure~(OF1) and per-Category F1-measure~(CF1) are used as the metrics to evaluate MLR performance. In particular, we find that the OpenImages test set is not fully labeled, resulting in a higher real class precision than the calculated class precision.

\subsection{Evaluation on Proposed Methods} 
\label{sec:proposed_methods}

\noindent
\textbf{Qualitative Results.} We report qualitative results in Figure~\ref{fig:visualization}. It shows that IDEA can additionally recognize image tags which do not exist in the original co-occurrent texts. These tags are entirely learned from the texts and recognized online, including objects~(e.g., \textit{"boat"}), scenes~(e.g., \textit{"snow"}), attributes~(e.g., \textit{"stained"}), actions~(e.g., \textit{"swimming"}), etc. More example visualization results are provided in Appendix~\ref{sec:more_visualization}.

\vspace{5pt}
\noindent
\textbf{Quantitative Results.} In Table~\ref{tab:proposed_methods}, we ablate our IDEA to demonstrate its effectiveness, mainly including two segments, Image-Text Contrastive learning~(ITC) and Multi-Label Recognition~(MLR). Only training ITC consistent with CLIP is the baseline for comparison. 
When ITC is combined with MLR, both zero-shot accuracy and multi-label performance have improvements. Specifically, the top-1 accuracy of zero-shot evaluation on ImageNet-1k~(IN1k) improves by 0.76\%, and the mAP of MLR on OpenImages improves from 30.89\% to 34.90\%. This indicates that although ITC and MLR are two distinct tasks, both can essentially improve the representation of the visual backbone. We consider the following works reflect the same finding: (i) CLIP~\cite{radford2021learning} can achieve superb performance only using visual backbone with linear-probe representation learning. (ii) ~\cite{li2021supervision,mu2021slip} have proposed that the combination of visual self-supervised learning and image-text contrastive learning can boost the model performance.

On this basis, leveraging Tag2Text to convert the additionally recognized tags by MLR into texts substantially improves the zero-shot performance. Specifically, compared with CLIP, IDEA has 2.18\% and 1.02\% improvements on the top-1 accuracy of IN1k zero-shot evaluation with COCO\&VG and CC-3M pre-training, respectively. We consider that cleaning the noise in the tag list~(e.g., \textit{polysemy, synonyms}) can further promote the model performance.


\subsection{Additional Experimental Comparison}
\vspace{5pt}
\noindent
\textbf{Detailed Zero-Shot Comparison.} Table ~\ref{tab:zero-shot} shows the detailed zero-shot comparison between CLIP and IDEA, demonstrating the generalization of IDEA. When pre-training on COCO\&VG dataset, IDEA has improvements on 7 out of 9 datasets, with an average increase of 0.62\% on top-1 accuracy. In addition, IDEA also achieves 0.63\% average top-1 accuracy improvement on 7 out of 9 datasets when pre-training on larger CC-3M dataset. We check the the categories of DTD and Food101 datasets which are fine-grained texture categories and food categories. We consider that the additional coarse-grained categories identified by IDEA negatively affect the performance on these datasets and increasing the fine-grained categories of the tag list can mitigate this effect. 


\begin{table*}[ht]
\setlength{\tabcolsep}{2.5mm}{
\centering
\caption{Detailed Top-1 zero-shot evaluation on 9 different downstream datasets.}
\vspace{-0.5em}
\label{tab:zero-shot}
\begin{tabular}{c|ccccccccc|c}
 Method& \rotatebox{90}{CIFAR10} & \rotatebox{90}{CIFAR100} & \rotatebox{90}{Caltech101}  & \rotatebox{90}{Flowers102} & \rotatebox{90}{Food101} & \rotatebox{90}{SUN397} & \rotatebox{90}{DTD}  & \rotatebox{90}{OxfordPets} &  \rotatebox{90}{\textbf{ImageNet}} & \rotatebox{90}{\textbf{Average}} \\ \shline
CLIP   & \textbf{69.56} &24.81 &10.04  &2.05  &13.27  &19.77  &\textbf{9.95}   &6.60  & 14.54  &  18.95\\
\textbf{IDEA}  &68.53  &\textbf{27.62} &\textbf{11.05}  &\textbf{2.19}  &\textbf{13.90}  &\textbf{20.28}  &8.71    &\textbf{7.09}  & \textbf{16.72}  &  \textbf{19.57}\\ \hline
CLIP~(3M) &84.80  &54.97  &15.94   &13.34  & \textbf{19.63} &42.06  &\textbf{22.87}  & 18.29   & 35.30 &34.13  \\
\textbf{IDEA~(3M)} &\textbf{84.94}  & \textbf{56.91} &\textbf{16.00}   &\textbf{16.09}  & 18.49 &\textbf{42.20}  &22.08  & \textbf{19.79}  & \textbf{36.32} & \textbf{34.76}
\end{tabular}}
\end{table*}

\vspace{5pt}
\noindent
\textbf{Ablation Study.} We study the impact of various improvements on MLR in Table~\ref{tab:mlr_ablation_study}. The recognition head with CLS refers to directly utilize visual global feature CLS followed by a fully connected layer like single-label classification. When the classification head and classification loss are developed to ML-Decoder and SPLC, both help improve performance.

\begin{table}[h]
\caption{Ablation study on MLR. CLS: utilizs visual global feature CLS followed by a fully connected layer.}
\label{tab:mlr_ablation_study}
\begin{tabular}{cc|cc|c}
\multicolumn{2}{c|}{Recognition Head} & \multicolumn{2}{c|}{Recognition Loss} &     \\
CLS            & ML-Decoder              & BCE        & SPLC        & mAP \\ \shline
\ding{51}      &                      &   \ding{51}             &                 & 26.89    \\
               &    \ding{51}          &  \ding{51}              &                 & 28.70    \\ \hline
               &     \ding{51}       &     &  \ding{51}       &  30.89 \\
\end{tabular}
\vspace{-0.5em}
\end{table}

\label{label2text_ablation_study}
We study the effect of various operations on Tag2Text in Table~\ref{tab:label2text_ablation_study}. Directly adopting Tag2Text degrades the zero-shot performance as most recognized tags are high frequency categories~(e.g., \textit{"people"}). RmTopFreq deletes these categories in the tag list, which increases IN1k zero-shot top-1 accuracy by 2.30\%. This phenomenon is consistent with~\cite{tejankar2021fistful} that removing the frequent words in BoW can make the ITC task harder and boost the model performance. Besides, we find that retaining the extracted tags from the original text, which are converted with the recognized tags to Tag2Text, can further improve the performance. We speculate that in addition to increasing the frequency of uncommon categories, it also makes the new text semantics more plausible.

\begin{table}[h]
\caption{Ablation study on Tag2Text. RmTopFreq: removes top-t most frequent tags. Retain: retains the original tags into Tag2Text.}
\label{tab:label2text_ablation_study}
\begin{tabular}{l|cc|cc}
           &\multicolumn{2}{c|}{IN1k Zero-Shot} & \multicolumn{2}{c}{Average Zero-Shot}   \\
Method     &  Top-1         & Top-5          &  Top-1         & Top-5\\ \shline
ITC    &   14.54       & 31.18  &  18.95  &35.61\\ 
+MLR    &   15.30       & 31.43  &  18.91 &36.60\\ \hline
+Tag2Text&    13.96       &30.41  & 18.63 &36.12 \\ 
\ \ +RmTopFreq &    16.26       &   33.30    & 19.06 & 37.25 \\
\ \ \ \ +Retain    &   16.72      & 34.04      &19.57   &37.70

\end{tabular}
\end{table}

\vspace{5pt}
\noindent
\textbf{Linear Probe Evaluation.} Following~\cite{radford2021learning}, we make an additional ImageNet-1k linear probe evaluation, which is commonly used in the field of self-supervised learning to evaluate the image representation. Linear probe refers to the backbone of the image encoder being fixed, and only one additional classifier is trained.
We follow the hyperparameters setting of DINO~\cite{caron2021emerging} for linear probe evaluation of the vision transformer. It is noteworthy that we train CLIP and IDEA from scratch without initializing the image encoder with ImageNet weights for fair comparison in this section.
\begin{table}[h]
\caption{Comparison results of ImageNet-1k linear probe evaluation.}
\label{tab:with BLIP}
\begin{tabular}{c|c|c}
Method & Top-1 Accuracy         & Top-5  Accuracy \\ \shline
CLIP   & 30.54                  & 53.12                \\ \hline
\textbf{IDEA*}   &  \textbf{31.63} & \textbf{54.32}                \\ \hline
\textbf{IDEA}  &  \textbf{31.41} & \textbf{54.41}                \\ 
\end{tabular}
\end{table}

\noindent
\textbf{Combination with BLIP.} From the perspective of data, our IDEA does not conflict with BLIP~\cite{li2022blip} and can be combined with BLIP. We adopt the released CC-3M dataset with BLIP CapFilt augment. As shown in Table~\ref{tab:with BLIP}, BLIP can provide more high-quality sentences and more precise tags for Multi-Label Recognition in IDEA, which can further boost the performance. 

\begin{table}[h]
\caption{Comparison results of ImageNet-1k zero-shot evaluation with different pre-traing datasets.}
\label{tab:with BLIP}
\begin{tabular}{c|c|c}
Method & Dataset           & IN1k Zero-Shot \\ \shline
CLIP   & CC-3M                   & 35.30                \\ \hline
 \textbf{IDEA}   & CC-3M                   & \textbf{36.32}                \\ \hline
CLIP   & CC-3M with BLIP augment & 37.40                \\ \hline
 \textbf{IDEA}  & CC-3M with BLIP augment & \textbf{39.36}               
\end{tabular}
\end{table}


\subsection{Further Analysis}
\noindent\textbf{Computational Cost.} 
Simple algorithm, low complexity, and low manual annotation cost are the core of scaling up supervision. From the perspective of training efficiency, our work focuses on boosting the performance of fusion encoder VLP models~(e.g. CLIP~\cite{radford2021learning}, ALGIN~\cite{jia2021scaling}) with a small extra computational cost, thus it can scale up large-scale datasets with high training efficiency. 
Compared with CLIP, IDEA only additionally adopts a recognition head to accomplish MLR in which FLOPs only increase by \textbf{3.88\%~(22.42 GFLOPs vs. 23.29 GFLOPs)}. According to an arbitrarily extended tag list, the image tag information is entirely extracted from the texts without manual annotation. As a comparison, the FLOPs of Faster RCNN-FPN is approximately 180~GFLOPs~\cite{carion2020end}, and labeling bounding boxes can be expensive.

\begin{figure}[h]
  \centering
  \includegraphics[width=.45\textwidth]{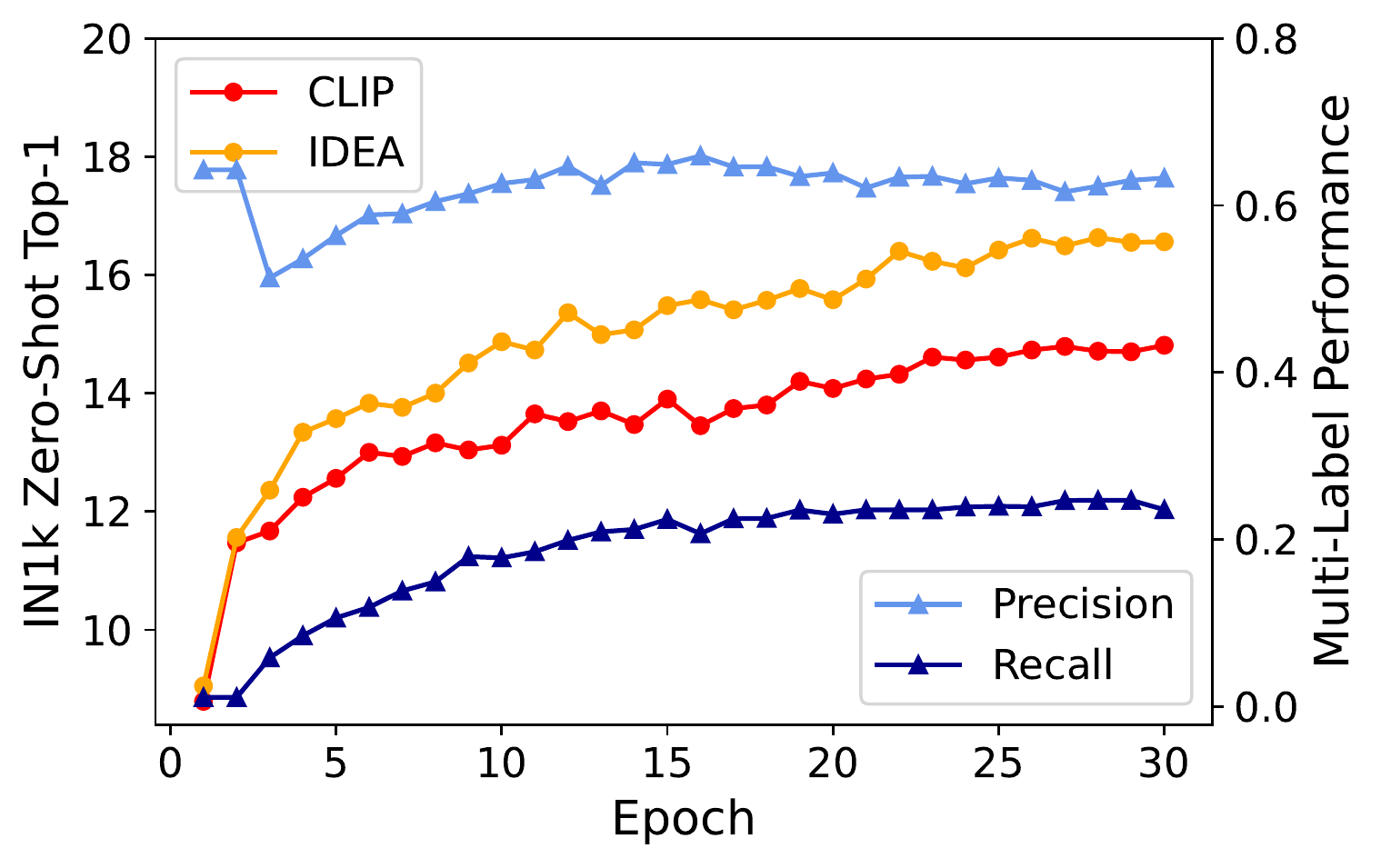} 
  \vspace{-0.5em}
  \caption{Performance improvement analysis. $\bullet$ lines with the left axis: the results of zero-shot evaluation on ImageNet-1k with different epochs. $\blacktriangle$ lines with the right axis: the precision \& recall of tags identified online.}
  \vspace{-0.8em}
  \label{fig:im1k-epoch}
\end{figure}

\vspace{5pt}
\noindent
\textbf{Performance Boost Analysis.} To further analyze how IDEA improves the performance, we plot the IN1k zero-shot results and multi-label performance through different training epochs~(see Figure~\ref{fig:im1k-epoch}). By adopting SPLC, IDEA can increasingly recognize missing tags online~(gradually increasing recall) while ensuring high precision. As illustrated in Figure~\ref{fig:im1k-epoch}, IDEA obviously outperforms CLIP with more missing tags identified.

\begin{figure}[h]
  \centering
  	\subfigure{
    	\begin{minipage}[t]{0.4\linewidth}
    	\centering
    	\includegraphics[width=1.25in]{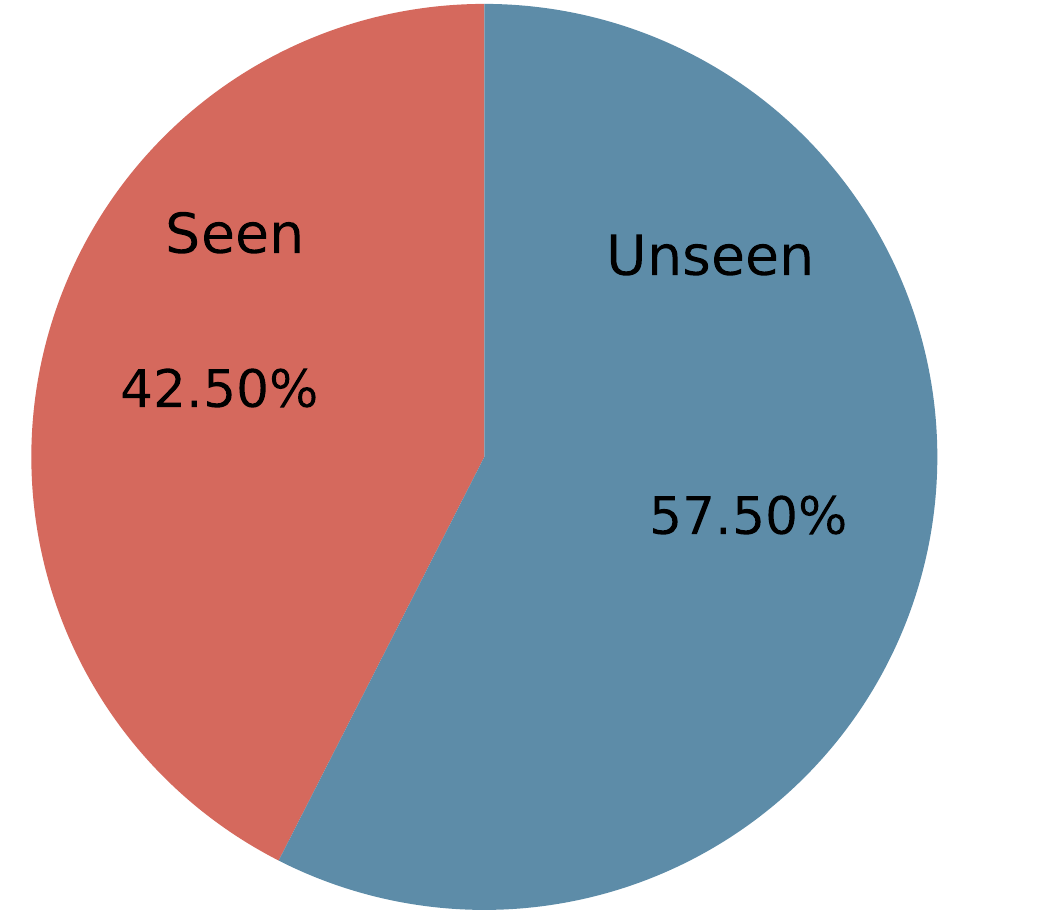}
    	
    	\scriptsize (a)
	    \end{minipage}%
	}%
	  	\subfigure{
    	\begin{minipage}[t]{0.5\linewidth}
    	\centering
    	\includegraphics[width=1.95in]{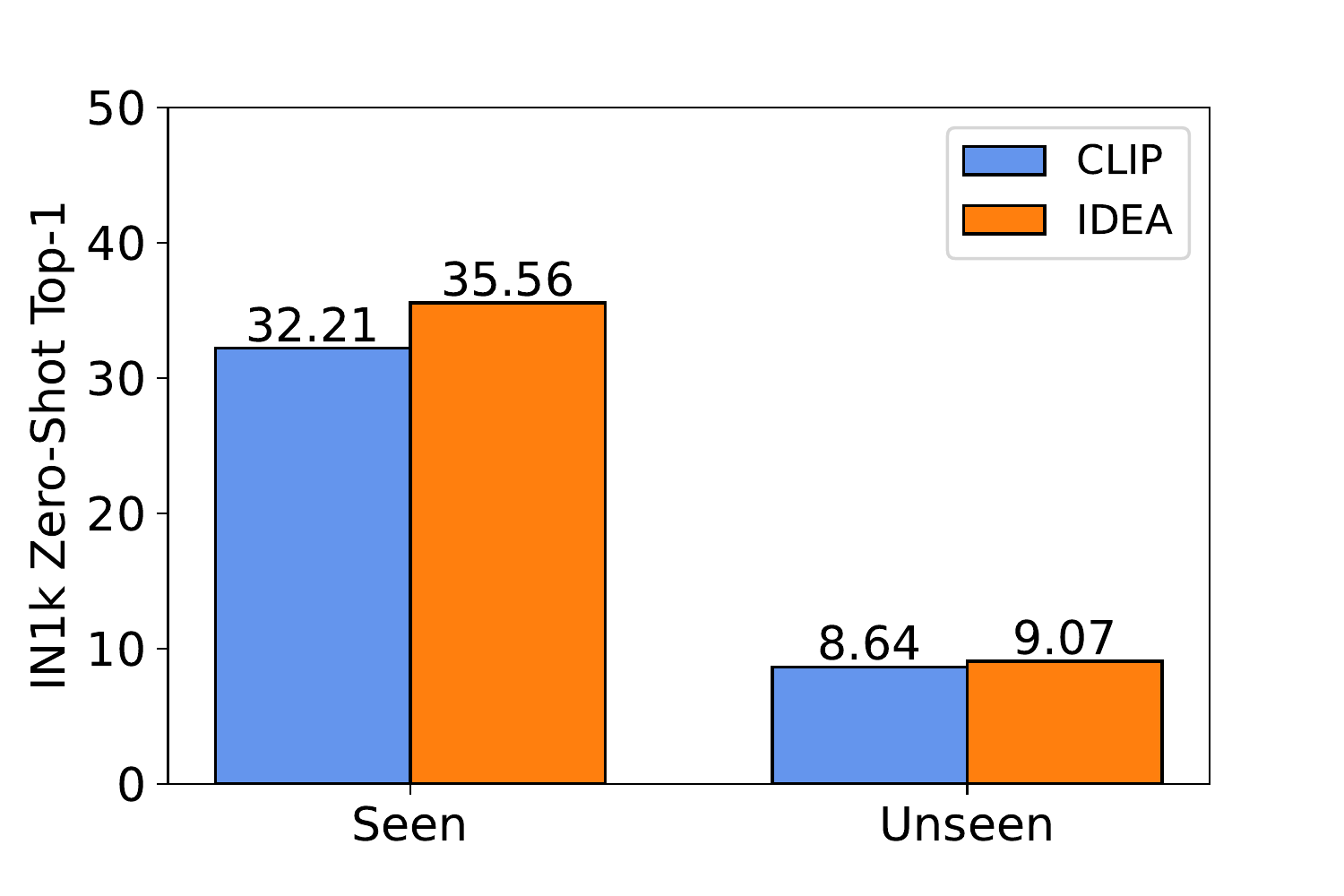}
    	\scriptsize (b)
	    \end{minipage}%
	}%
    \vspace{-0.5em}
  \caption{Illustration of categories overlap. (a)~The proportion of seen and unseen categories in the IN1k zero-shot evaluation promoted by IDEA.  (b)~The average performance comparison of seen and unseen categories in IN1k zero-shot evaluation. }
  \vspace{-0.5em}
  \label{fig:overlap_analysis}
\end{figure}

\noindent
\textbf{Categories Overlap.} 
\label{sec:overlap_analysis}
Since partial categories of downstream datasets overlap with the tag list, we analyze which part is mainly improved. We define the categories that overlap with the tag list as "seen" and the categories that do not overlap as "unseen". As shown in Figure~\ref{fig:overlap_analysis}(a), we examine all categories in the IN1k zero-shot evaluation promoted by IDEA and discover 57.50\% of them are unseen. This illustrates that although unseen are not included in the tag list, they can be indirectly corrected by additional supervision provided by IDEA. Figure \ref{fig:overlap_analysis}(b) shows the
average performance comparison of seen and unseen categories, respectively. Seen categories achieve a remarkable performance boost that the average top-1 accuracy is improved by 3.35\%, while the performance of the unseen categories does not decline and improves to some extent.

\begin{figure}[ht]
  \centering
  \includegraphics[width=.45\textwidth]{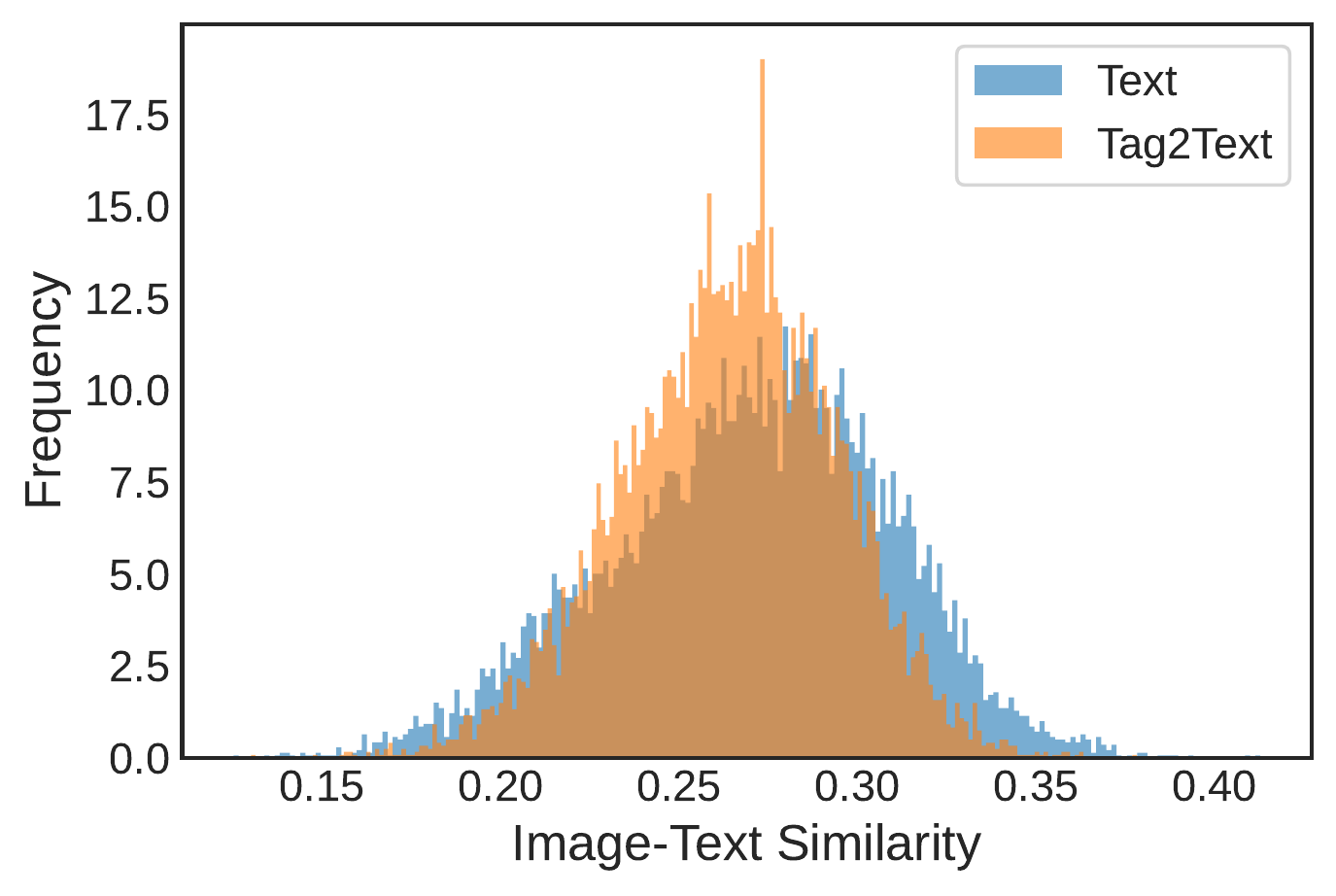} 
  \vspace{-0.5em}
  \caption{Distribution of image-text similarity calculated by the official CLIP model. Despite the words of Tag2Text are disordered, Tag2Text also has high similarity with the corresponding image compared with the original corresponding text.}
  \vspace{-0.5em}
  \label{fig:similarity}
\end{figure}

\vspace{5pt}
\noindent
\textbf{Text vs. Tag2Text.} We analyze whether Tag2Text with cluttered words can provide meaningful textual supervised information. The officially released CLIP model~\cite{radford2021learning} is pre-trained on 400M image-text pairs, which has a strong ability to judge the matching degree of image-text pairs. Therefore, we compare the image-text similarity between the original co-occurrent text and Tag2Text by utilizing the official CLIP ViT-B/32 model~(see Figure~\ref{fig:similarity}). On the one hand, since the words of Tag2Text are disordered, it is not easy to obtain extremely high image-text similarity. On the other hand, benefiting from the high accuracy of the identified tags, almost all Tag2Texts have a high image-text similarity, indicating that Tag2Text can provide practical and effective textual supervision.

\section{Conclusion}

This paper proposes IDEA to improve visual representation learning by enriching supervisory signals with textual information. IDEA can provide more explicit supervision online, including multiple valuable tags and texts composed by multiple tags. Extensive experiments have demonstrated IDEA can significantly boost performance with a small extra computational cost.

Some areas that we do not explore in depth are as follows: further cleaning the tag list to extract higher-quality tags, better utilization of identified tag information, extracting phrase information in addition to tag information, etc. IDEA shows exploiting the fine-grained image information implied in texts is a valuable prospect, and we hope to provide some inspiration for other works.

\begin{acks}
This work was supported by National Natural Science Foundation of China (No.62172101, No.61976057); the National Key Research and Development Program of China (No.2021ZD0113501); the Science and Technology Major Project of Commission of Science and Technology of Shanghai (No.20511100800, No.21511104502, No.21511100500).
\end{acks}




\bibliographystyle{ACM-Reference-Format}
\balance
\bibliography{mm2022}


\begin{thebibliography}{51}


\ifx \showCODEN    \undefined \def \showCODEN     #1{\unskip}     \fi
\ifx \showDOI      \undefined \def \showDOI       #1{#1}\fi
\ifx \showISBNx    \undefined \def \showISBNx     #1{\unskip}     \fi
\ifx \showISBNxiii \undefined \def \showISBNxiii  #1{\unskip}     \fi
\ifx \showISSN     \undefined \def \showISSN      #1{\unskip}     \fi
\ifx \showLCCN     \undefined \def \showLCCN      #1{\unskip}     \fi
\ifx \shownote     \undefined \def \shownote      #1{#1}          \fi
\ifx \showarticletitle \undefined \def \showarticletitle #1{#1}   \fi
\ifx \showURL      \undefined \def \showURL       {\relax}        \fi
\providecommand\bibfield[2]{#2}
\providecommand\bibinfo[2]{#2}
\providecommand\natexlab[1]{#1}
\providecommand\showeprint[2][]{arXiv:#2}

\bibitem[\protect\citeauthoryear{Bossard, Guillaumin, and Gool}{Bossard
  et~al\mbox{.}}{2014}]%
        {bossard2014food}
\bibfield{author}{\bibinfo{person}{Lukas Bossard}, \bibinfo{person}{Matthieu
  Guillaumin}, {and} \bibinfo{person}{Luc~Van Gool}.}
  \bibinfo{year}{2014}\natexlab{}.
\newblock \showarticletitle{Food-101--mining discriminative components with
  random forests}. In \bibinfo{booktitle}{\emph{European conference on computer
  vision}}. Springer, \bibinfo{pages}{446--461}.
\newblock


\bibitem[\protect\citeauthoryear{Carion, Massa, Synnaeve, Usunier, Kirillov,
  and Zagoruyko}{Carion et~al\mbox{.}}{2020}]%
        {carion2020end}
\bibfield{author}{\bibinfo{person}{Nicolas Carion}, \bibinfo{person}{Francisco
  Massa}, \bibinfo{person}{Gabriel Synnaeve}, \bibinfo{person}{Nicolas
  Usunier}, \bibinfo{person}{Alexander Kirillov}, {and} \bibinfo{person}{Sergey
  Zagoruyko}.} \bibinfo{year}{2020}\natexlab{}.
\newblock \showarticletitle{End-to-end object detection with transformers}. In
  \bibinfo{booktitle}{\emph{European conference on computer vision}}. Springer,
  \bibinfo{pages}{213--229}.
\newblock


\bibitem[\protect\citeauthoryear{Caron, Touvron, Misra, J{\'e}gou, Mairal,
  Bojanowski, and Joulin}{Caron et~al\mbox{.}}{2021}]%
        {caron2021emerging}
\bibfield{author}{\bibinfo{person}{Mathilde Caron}, \bibinfo{person}{Hugo
  Touvron}, \bibinfo{person}{Ishan Misra}, \bibinfo{person}{Herv{\'e}
  J{\'e}gou}, \bibinfo{person}{Julien Mairal}, \bibinfo{person}{Piotr
  Bojanowski}, {and} \bibinfo{person}{Armand Joulin}.}
  \bibinfo{year}{2021}\natexlab{}.
\newblock \showarticletitle{Emerging properties in self-supervised vision
  transformers}. In \bibinfo{booktitle}{\emph{Proceedings of the IEEE/CVF
  International Conference on Computer Vision}}. \bibinfo{pages}{9650--9660}.
\newblock


\bibitem[\protect\citeauthoryear{Chen, Fang, Lin, Vedantam, Gupta, Doll{\'a}r,
  and Zitnick}{Chen et~al\mbox{.}}{2015}]%
        {chen2015microsoft}
\bibfield{author}{\bibinfo{person}{Xinlei Chen}, \bibinfo{person}{Hao Fang},
  \bibinfo{person}{Tsung-Yi Lin}, \bibinfo{person}{Ramakrishna Vedantam},
  \bibinfo{person}{Saurabh Gupta}, \bibinfo{person}{Piotr Doll{\'a}r}, {and}
  \bibinfo{person}{C~Lawrence Zitnick}.} \bibinfo{year}{2015}\natexlab{}.
\newblock \showarticletitle{Microsoft coco captions: Data collection and
  evaluation server}.
\newblock \bibinfo{journal}{\emph{arXiv preprint arXiv:1504.00325}}
  (\bibinfo{year}{2015}).
\newblock


\bibitem[\protect\citeauthoryear{Chen, Li, Yu, El~Kholy, Ahmed, Gan, Cheng, and
  Liu}{Chen et~al\mbox{.}}{2020}]%
        {chen2020uniter}
\bibfield{author}{\bibinfo{person}{Yen-Chun Chen}, \bibinfo{person}{Linjie Li},
  \bibinfo{person}{Licheng Yu}, \bibinfo{person}{Ahmed El~Kholy},
  \bibinfo{person}{Faisal Ahmed}, \bibinfo{person}{Zhe Gan},
  \bibinfo{person}{Yu Cheng}, {and} \bibinfo{person}{Jingjing Liu}.}
  \bibinfo{year}{2020}\natexlab{}.
\newblock \showarticletitle{Uniter: Universal image-text representation
  learning}. In \bibinfo{booktitle}{\emph{European conference on computer
  vision}}. Springer, \bibinfo{pages}{104--120}.
\newblock


\bibitem[\protect\citeauthoryear{Cimpoi, Maji, Kokkinos, Mohamed, and
  Vedaldi}{Cimpoi et~al\mbox{.}}{2014}]%
        {cimpoi2014describing}
\bibfield{author}{\bibinfo{person}{Mircea Cimpoi}, \bibinfo{person}{Subhransu
  Maji}, \bibinfo{person}{Iasonas Kokkinos}, \bibinfo{person}{Sammy Mohamed},
  {and} \bibinfo{person}{Andrea Vedaldi}.} \bibinfo{year}{2014}\natexlab{}.
\newblock \showarticletitle{Describing textures in the wild}. In
  \bibinfo{booktitle}{\emph{Proceedings of the IEEE conference on computer
  vision and pattern recognition}}. \bibinfo{pages}{3606--3613}.
\newblock


\bibitem[\protect\citeauthoryear{Cole, Mac~Aodha, Lorieul, Perona, Morris, and
  Jojic}{Cole et~al\mbox{.}}{2021}]%
        {cole2021multi}
\bibfield{author}{\bibinfo{person}{Elijah Cole}, \bibinfo{person}{Oisin
  Mac~Aodha}, \bibinfo{person}{Titouan Lorieul}, \bibinfo{person}{Pietro
  Perona}, \bibinfo{person}{Dan Morris}, {and} \bibinfo{person}{Nebojsa
  Jojic}.} \bibinfo{year}{2021}\natexlab{}.
\newblock \showarticletitle{Multi-Label Learning from Single Positive Labels}.
  In \bibinfo{booktitle}{\emph{Proceedings of the IEEE/CVF Conference on
  Computer Vision and Pattern Recognition}}. \bibinfo{pages}{933--942}.
\newblock


\bibitem[\protect\citeauthoryear{Cubuk, Zoph, Shlens, and Le}{Cubuk
  et~al\mbox{.}}{2020}]%
        {cubuk2020randaugment}
\bibfield{author}{\bibinfo{person}{Ekin~D Cubuk}, \bibinfo{person}{Barret
  Zoph}, \bibinfo{person}{Jonathon Shlens}, {and} \bibinfo{person}{Quoc~V Le}.}
  \bibinfo{year}{2020}\natexlab{}.
\newblock \showarticletitle{Randaugment: Practical automated data augmentation
  with a reduced search space}. In \bibinfo{booktitle}{\emph{Proceedings of the
  IEEE/CVF Conference on Computer Vision and Pattern Recognition Workshops}}.
  \bibinfo{pages}{702--703}.
\newblock


\bibitem[\protect\citeauthoryear{Deng, Dong, Socher, Li, Li, and Fei-Fei}{Deng
  et~al\mbox{.}}{2009}]%
        {deng2009imagenet}
\bibfield{author}{\bibinfo{person}{Jia Deng}, \bibinfo{person}{Wei Dong},
  \bibinfo{person}{Richard Socher}, \bibinfo{person}{Li-Jia Li},
  \bibinfo{person}{Kai Li}, {and} \bibinfo{person}{Li Fei-Fei}.}
  \bibinfo{year}{2009}\natexlab{}.
\newblock \showarticletitle{Imagenet: A large-scale hierarchical image
  database}. In \bibinfo{booktitle}{\emph{2009 IEEE conference on computer
  vision and pattern recognition}}. Ieee, \bibinfo{pages}{248--255}.
\newblock


\bibitem[\protect\citeauthoryear{Devlin, Chang, Lee, and Toutanova}{Devlin
  et~al\mbox{.}}{2018}]%
        {devlin2018bert}
\bibfield{author}{\bibinfo{person}{Jacob Devlin}, \bibinfo{person}{Ming-Wei
  Chang}, \bibinfo{person}{Kenton Lee}, {and} \bibinfo{person}{Kristina
  Toutanova}.} \bibinfo{year}{2018}\natexlab{}.
\newblock \showarticletitle{Bert: Pre-training of deep bidirectional
  transformers for language understanding}.
\newblock \bibinfo{journal}{\emph{arXiv preprint arXiv:1810.04805}}
  (\bibinfo{year}{2018}).
\newblock


\bibitem[\protect\citeauthoryear{Dosovitskiy, Beyer, Kolesnikov, Weissenborn,
  Zhai, Unterthiner, Dehghani, Minderer, Heigold, Gelly,
  et~al\mbox{.}}{Dosovitskiy et~al\mbox{.}}{2020}]%
        {dosovitskiy2020image}
\bibfield{author}{\bibinfo{person}{Alexey Dosovitskiy}, \bibinfo{person}{Lucas
  Beyer}, \bibinfo{person}{Alexander Kolesnikov}, \bibinfo{person}{Dirk
  Weissenborn}, \bibinfo{person}{Xiaohua Zhai}, \bibinfo{person}{Thomas
  Unterthiner}, \bibinfo{person}{Mostafa Dehghani}, \bibinfo{person}{Matthias
  Minderer}, \bibinfo{person}{Georg Heigold}, \bibinfo{person}{Sylvain Gelly},
  {et~al\mbox{.}}} \bibinfo{year}{2020}\natexlab{}.
\newblock \showarticletitle{An image is worth 16x16 words: Transformers for
  image recognition at scale}.
\newblock \bibinfo{journal}{\emph{arXiv preprint arXiv:2010.11929}}
  (\bibinfo{year}{2020}).
\newblock


\bibitem[\protect\citeauthoryear{Dou, Xu, Gan, Wang, Wang, Wang, Zhu, Liu,
  Zeng, et~al\mbox{.}}{Dou et~al\mbox{.}}{2021}]%
        {dou2021empirical}
\bibfield{author}{\bibinfo{person}{Zi-Yi Dou}, \bibinfo{person}{Yichong Xu},
  \bibinfo{person}{Zhe Gan}, \bibinfo{person}{Jianfeng Wang},
  \bibinfo{person}{Shuohang Wang}, \bibinfo{person}{Lijuan Wang},
  \bibinfo{person}{Chenguang Zhu}, \bibinfo{person}{Zicheng Liu},
  \bibinfo{person}{Michael Zeng}, {et~al\mbox{.}}}
  \bibinfo{year}{2021}\natexlab{}.
\newblock \showarticletitle{An Empirical Study of Training End-to-End
  Vision-and-Language Transformers}.
\newblock \bibinfo{journal}{\emph{arXiv preprint arXiv:2111.02387}}
  (\bibinfo{year}{2021}).
\newblock


\bibitem[\protect\citeauthoryear{Du, Liu, Li, and Zhao}{Du
  et~al\mbox{.}}{2022}]%
        {du2022survey}
\bibfield{author}{\bibinfo{person}{Yifan Du}, \bibinfo{person}{Zikang Liu},
  \bibinfo{person}{Junyi Li}, {and} \bibinfo{person}{Wayne~Xin Zhao}.}
  \bibinfo{year}{2022}\natexlab{}.
\newblock \showarticletitle{A survey of vision-language pre-trained models}.
\newblock \bibinfo{journal}{\emph{arXiv preprint arXiv:2202.10936}}
  (\bibinfo{year}{2022}).
\newblock


\bibitem[\protect\citeauthoryear{Durand, Mehrasa, and Mori}{Durand
  et~al\mbox{.}}{2019}]%
        {durand2019learning}
\bibfield{author}{\bibinfo{person}{Thibaut Durand}, \bibinfo{person}{Nazanin
  Mehrasa}, {and} \bibinfo{person}{Greg Mori}.}
  \bibinfo{year}{2019}\natexlab{}.
\newblock \showarticletitle{Learning a deep convnet for multi-label
  classification with partial labels}. In \bibinfo{booktitle}{\emph{Proceedings
  of the IEEE/CVF Conference on Computer Vision and Pattern Recognition}}.
  \bibinfo{pages}{647--657}.
\newblock


\bibitem[\protect\citeauthoryear{Everingham and Winn}{Everingham and
  Winn}{2011}]%
        {everingham2011pascal}
\bibfield{author}{\bibinfo{person}{Mark Everingham} {and} \bibinfo{person}{John
  Winn}.} \bibinfo{year}{2011}\natexlab{}.
\newblock \showarticletitle{The pascal visual object classes challenge 2012
  (voc2012) development kit}.
\newblock \bibinfo{journal}{\emph{Pattern Analysis, Statistical Modelling and
  Computational Learning, Tech. Rep}}  \bibinfo{volume}{8}
  (\bibinfo{year}{2011}), \bibinfo{pages}{5}.
\newblock


\bibitem[\protect\citeauthoryear{Fang, Xiong, Xu, and Chen}{Fang
  et~al\mbox{.}}{2021}]%
        {fang2021clip2video}
\bibfield{author}{\bibinfo{person}{Han Fang}, \bibinfo{person}{Pengfei Xiong},
  \bibinfo{person}{Luhui Xu}, {and} \bibinfo{person}{Yu Chen}.}
  \bibinfo{year}{2021}\natexlab{}.
\newblock \showarticletitle{Clip2video: Mastering video-text retrieval via
  image clip}.
\newblock \bibinfo{journal}{\emph{arXiv preprint arXiv:2106.11097}}
  (\bibinfo{year}{2021}).
\newblock


\bibitem[\protect\citeauthoryear{Fei-Fei, Fergus, and Perona}{Fei-Fei
  et~al\mbox{.}}{2004}]%
        {fei2004learning}
\bibfield{author}{\bibinfo{person}{Li Fei-Fei}, \bibinfo{person}{Rob Fergus},
  {and} \bibinfo{person}{Pietro Perona}.} \bibinfo{year}{2004}\natexlab{}.
\newblock \showarticletitle{Learning generative visual models from few training
  examples: An incremental bayesian approach tested on 101 object categories}.
  In \bibinfo{booktitle}{\emph{2004 conference on computer vision and pattern
  recognition workshop}}. IEEE, \bibinfo{pages}{178--178}.
\newblock


\bibitem[\protect\citeauthoryear{Gu, Lin, Kuo, and Cui}{Gu
  et~al\mbox{.}}{2021}]%
        {gu2021zero}
\bibfield{author}{\bibinfo{person}{Xiuye Gu}, \bibinfo{person}{Tsung-Yi Lin},
  \bibinfo{person}{Weicheng Kuo}, {and} \bibinfo{person}{Yin Cui}.}
  \bibinfo{year}{2021}\natexlab{}.
\newblock \showarticletitle{Zero-shot detection via vision and language
  knowledge distillation}.
\newblock \bibinfo{journal}{\emph{arXiv e-prints}} (\bibinfo{year}{2021}),
  \bibinfo{pages}{arXiv--2104}.
\newblock


\bibitem[\protect\citeauthoryear{Huang, Zeng, Huang, Liu, Fu, and Fu}{Huang
  et~al\mbox{.}}{2021}]%
        {huang2021seeing}
\bibfield{author}{\bibinfo{person}{Zhicheng Huang}, \bibinfo{person}{Zhaoyang
  Zeng}, \bibinfo{person}{Yupan Huang}, \bibinfo{person}{Bei Liu},
  \bibinfo{person}{Dongmei Fu}, {and} \bibinfo{person}{Jianlong Fu}.}
  \bibinfo{year}{2021}\natexlab{}.
\newblock \showarticletitle{Seeing out of the box: End-to-end pre-training for
  vision-language representation learning}. In
  \bibinfo{booktitle}{\emph{Proceedings of the IEEE/CVF Conference on Computer
  Vision and Pattern Recognition}}. \bibinfo{pages}{12976--12985}.
\newblock


\bibitem[\protect\citeauthoryear{Jia, Yang, Xia, Chen, Parekh, Pham, Le, Sung,
  Li, and Duerig}{Jia et~al\mbox{.}}{2021}]%
        {jia2021scaling}
\bibfield{author}{\bibinfo{person}{Chao Jia}, \bibinfo{person}{Yinfei Yang},
  \bibinfo{person}{Ye Xia}, \bibinfo{person}{Yi-Ting Chen},
  \bibinfo{person}{Zarana Parekh}, \bibinfo{person}{Hieu Pham},
  \bibinfo{person}{Quoc Le}, \bibinfo{person}{Yun-Hsuan Sung},
  \bibinfo{person}{Zhen Li}, {and} \bibinfo{person}{Tom Duerig}.}
  \bibinfo{year}{2021}\natexlab{}.
\newblock \showarticletitle{Scaling up visual and vision-language
  representation learning with noisy text supervision}. In
  \bibinfo{booktitle}{\emph{International Conference on Machine Learning}}.
  PMLR, \bibinfo{pages}{4904--4916}.
\newblock


\bibitem[\protect\citeauthoryear{Ju, Han, Zheng, Zhang, and Xie}{Ju
  et~al\mbox{.}}{2021}]%
        {ju2021prompting}
\bibfield{author}{\bibinfo{person}{Chen Ju}, \bibinfo{person}{Tengda Han},
  \bibinfo{person}{Kunhao Zheng}, \bibinfo{person}{Ya Zhang}, {and}
  \bibinfo{person}{Weidi Xie}.} \bibinfo{year}{2021}\natexlab{}.
\newblock \showarticletitle{Prompting Visual-Language Models for Efficient
  Video Understanding}.
\newblock \bibinfo{journal}{\emph{arXiv preprint arXiv:2112.04478}}
  (\bibinfo{year}{2021}).
\newblock


\bibitem[\protect\citeauthoryear{Krishna, Zhu, Groth, Johnson, Hata, Kravitz,
  Chen, Kalantidis, Li, Shamma, et~al\mbox{.}}{Krishna et~al\mbox{.}}{2017}]%
        {krishna2017visual}
\bibfield{author}{\bibinfo{person}{Ranjay Krishna}, \bibinfo{person}{Yuke Zhu},
  \bibinfo{person}{Oliver Groth}, \bibinfo{person}{Justin Johnson},
  \bibinfo{person}{Kenji Hata}, \bibinfo{person}{Joshua Kravitz},
  \bibinfo{person}{Stephanie Chen}, \bibinfo{person}{Yannis Kalantidis},
  \bibinfo{person}{Li-Jia Li}, \bibinfo{person}{David~A Shamma},
  {et~al\mbox{.}}} \bibinfo{year}{2017}\natexlab{}.
\newblock \showarticletitle{Visual genome: Connecting language and vision using
  crowdsourced dense image annotations}.
\newblock \bibinfo{journal}{\emph{International journal of computer vision}}
  \bibinfo{volume}{123}, \bibinfo{number}{1} (\bibinfo{year}{2017}),
  \bibinfo{pages}{32--73}.
\newblock


\bibitem[\protect\citeauthoryear{Krizhevsky, Hinton, et~al\mbox{.}}{Krizhevsky
  et~al\mbox{.}}{2009}]%
        {krizhevsky2009learning}
\bibfield{author}{\bibinfo{person}{Alex Krizhevsky}, \bibinfo{person}{Geoffrey
  Hinton}, {et~al\mbox{.}}} \bibinfo{year}{2009}\natexlab{}.
\newblock \showarticletitle{Learning multiple layers of features from tiny
  images}.
\newblock  (\bibinfo{year}{2009}).
\newblock


\bibitem[\protect\citeauthoryear{Kuznetsova, Rom, Alldrin, Uijlings, Krasin,
  Pont-Tuset, Kamali, Popov, Malloci, Kolesnikov, et~al\mbox{.}}{Kuznetsova
  et~al\mbox{.}}{2020}]%
        {kuznetsova2020open}
\bibfield{author}{\bibinfo{person}{Alina Kuznetsova}, \bibinfo{person}{Hassan
  Rom}, \bibinfo{person}{Neil Alldrin}, \bibinfo{person}{Jasper Uijlings},
  \bibinfo{person}{Ivan Krasin}, \bibinfo{person}{Jordi Pont-Tuset},
  \bibinfo{person}{Shahab Kamali}, \bibinfo{person}{Stefan Popov},
  \bibinfo{person}{Matteo Malloci}, \bibinfo{person}{Alexander Kolesnikov},
  {et~al\mbox{.}}} \bibinfo{year}{2020}\natexlab{}.
\newblock \showarticletitle{The open images dataset v4}.
\newblock \bibinfo{journal}{\emph{International Journal of Computer Vision}}
  \bibinfo{volume}{128}, \bibinfo{number}{7} (\bibinfo{year}{2020}),
  \bibinfo{pages}{1956--1981}.
\newblock


\bibitem[\protect\citeauthoryear{Li, Zhang, Zhang, Liu, Guo, Ni, Zhang, and
  Zhang}{Li et~al\mbox{.}}{2022b}]%
        {li2022vision}
\bibfield{author}{\bibinfo{person}{Feng Li}, \bibinfo{person}{Hao Zhang},
  \bibinfo{person}{Yi-Fan Zhang}, \bibinfo{person}{Shilong Liu},
  \bibinfo{person}{Jian Guo}, \bibinfo{person}{Lionel~M Ni},
  \bibinfo{person}{PengChuan Zhang}, {and} \bibinfo{person}{Lei Zhang}.}
  \bibinfo{year}{2022}\natexlab{b}.
\newblock \showarticletitle{Vision-Language Intelligence: Tasks, Representation
  Learning, and Large Models}.
\newblock \bibinfo{journal}{\emph{arXiv preprint arXiv:2203.01922}}
  (\bibinfo{year}{2022}).
\newblock


\bibitem[\protect\citeauthoryear{Li, Li, Xiong, and Hoi}{Li
  et~al\mbox{.}}{2022a}]%
        {li2022blip}
\bibfield{author}{\bibinfo{person}{Junnan Li}, \bibinfo{person}{Dongxu Li},
  \bibinfo{person}{Caiming Xiong}, {and} \bibinfo{person}{Steven Hoi}.}
  \bibinfo{year}{2022}\natexlab{a}.
\newblock \showarticletitle{BLIP: Bootstrapping Language-Image Pre-training for
  Unified Vision-Language Understanding and Generation}.
\newblock \bibinfo{journal}{\emph{arXiv preprint arXiv:2201.12086}}
  (\bibinfo{year}{2022}).
\newblock


\bibitem[\protect\citeauthoryear{Li, Selvaraju, Gotmare, Joty, Xiong, and
  Hoi}{Li et~al\mbox{.}}{2021b}]%
        {li2021align}
\bibfield{author}{\bibinfo{person}{Junnan Li}, \bibinfo{person}{Ramprasaath
  Selvaraju}, \bibinfo{person}{Akhilesh Gotmare}, \bibinfo{person}{Shafiq
  Joty}, \bibinfo{person}{Caiming Xiong}, {and} \bibinfo{person}{Steven
  Chu~Hong Hoi}.} \bibinfo{year}{2021}\natexlab{b}.
\newblock \showarticletitle{Align before fuse: Vision and language
  representation learning with momentum distillation}.
\newblock \bibinfo{journal}{\emph{Advances in Neural Information Processing
  Systems}}  \bibinfo{volume}{34} (\bibinfo{year}{2021}).
\newblock


\bibitem[\protect\citeauthoryear{Li, Yatskar, Yin, Hsieh, and Chang}{Li
  et~al\mbox{.}}{2019}]%
        {li2019visualbert}
\bibfield{author}{\bibinfo{person}{Liunian~Harold Li}, \bibinfo{person}{Mark
  Yatskar}, \bibinfo{person}{Da Yin}, \bibinfo{person}{Cho-Jui Hsieh}, {and}
  \bibinfo{person}{Kai-Wei Chang}.} \bibinfo{year}{2019}\natexlab{}.
\newblock \showarticletitle{Visualbert: A simple and performant baseline for
  vision and language}.
\newblock \bibinfo{journal}{\emph{arXiv preprint arXiv:1908.03557}}
  (\bibinfo{year}{2019}).
\newblock


\bibitem[\protect\citeauthoryear{Li, Yin, Li, Zhang, Hu, Zhang, Wang, Hu, Dong,
  Wei, et~al\mbox{.}}{Li et~al\mbox{.}}{2020}]%
        {li2020oscar}
\bibfield{author}{\bibinfo{person}{Xiujun Li}, \bibinfo{person}{Xi Yin},
  \bibinfo{person}{Chunyuan Li}, \bibinfo{person}{Pengchuan Zhang},
  \bibinfo{person}{Xiaowei Hu}, \bibinfo{person}{Lei Zhang},
  \bibinfo{person}{Lijuan Wang}, \bibinfo{person}{Houdong Hu},
  \bibinfo{person}{Li Dong}, \bibinfo{person}{Furu Wei}, {et~al\mbox{.}}}
  \bibinfo{year}{2020}\natexlab{}.
\newblock \showarticletitle{Oscar: Object-semantics aligned pre-training for
  vision-language tasks}. In \bibinfo{booktitle}{\emph{European Conference on
  Computer Vision}}. Springer, \bibinfo{pages}{121--137}.
\newblock


\bibitem[\protect\citeauthoryear{Li, Liang, Zhao, Cui, Ouyang, Shao, Yu, and
  Yan}{Li et~al\mbox{.}}{2021a}]%
        {li2021supervision}
\bibfield{author}{\bibinfo{person}{Yangguang Li}, \bibinfo{person}{Feng Liang},
  \bibinfo{person}{Lichen Zhao}, \bibinfo{person}{Yufeng Cui},
  \bibinfo{person}{Wanli Ouyang}, \bibinfo{person}{Jing Shao},
  \bibinfo{person}{Fengwei Yu}, {and} \bibinfo{person}{Junjie Yan}.}
  \bibinfo{year}{2021}\natexlab{a}.
\newblock \showarticletitle{Supervision exists everywhere: A data efficient
  contrastive language-image pre-training paradigm}.
\newblock \bibinfo{journal}{\emph{arXiv preprint arXiv:2110.05208}}
  (\bibinfo{year}{2021}).
\newblock


\bibitem[\protect\citeauthoryear{Liu, Zhang, Yang, Su, and Zhu}{Liu
  et~al\mbox{.}}{2021}]%
        {liu2021query2label}
\bibfield{author}{\bibinfo{person}{Shilong Liu}, \bibinfo{person}{Lei Zhang},
  \bibinfo{person}{Xiao Yang}, \bibinfo{person}{Hang Su}, {and}
  \bibinfo{person}{Jun Zhu}.} \bibinfo{year}{2021}\natexlab{}.
\newblock \showarticletitle{Query2label: A simple transformer way to
  multi-label classification}.
\newblock \bibinfo{journal}{\emph{arXiv preprint arXiv:2107.10834}}
  (\bibinfo{year}{2021}).
\newblock


\bibitem[\protect\citeauthoryear{Loper and Bird}{Loper and Bird}{2002}]%
        {loper2002nltk}
\bibfield{author}{\bibinfo{person}{Edward Loper} {and} \bibinfo{person}{Steven
  Bird}.} \bibinfo{year}{2002}\natexlab{}.
\newblock \showarticletitle{Nltk: The natural language toolkit}.
\newblock \bibinfo{journal}{\emph{arXiv preprint cs/0205028}}
  (\bibinfo{year}{2002}).
\newblock


\bibitem[\protect\citeauthoryear{Loshchilov and Hutter}{Loshchilov and
  Hutter}{2016}]%
        {loshchilov2016sgdr}
\bibfield{author}{\bibinfo{person}{Ilya Loshchilov} {and}
  \bibinfo{person}{Frank Hutter}.} \bibinfo{year}{2016}\natexlab{}.
\newblock \showarticletitle{Sgdr: Stochastic gradient descent with warm
  restarts}.
\newblock \bibinfo{journal}{\emph{arXiv preprint arXiv:1608.03983}}
  (\bibinfo{year}{2016}).
\newblock


\bibitem[\protect\citeauthoryear{Lu, Batra, Parikh, and Lee}{Lu
  et~al\mbox{.}}{2019}]%
        {lu2019vilbert}
\bibfield{author}{\bibinfo{person}{Jiasen Lu}, \bibinfo{person}{Dhruv Batra},
  \bibinfo{person}{Devi Parikh}, {and} \bibinfo{person}{Stefan Lee}.}
  \bibinfo{year}{2019}\natexlab{}.
\newblock \showarticletitle{Vilbert: Pretraining task-agnostic visiolinguistic
  representations for vision-and-language tasks}.
\newblock \bibinfo{journal}{\emph{Advances in neural information processing
  systems}}  \bibinfo{volume}{32} (\bibinfo{year}{2019}).
\newblock


\bibitem[\protect\citeauthoryear{Luo, Ji, Zhong, Chen, Lei, Duan, and Li}{Luo
  et~al\mbox{.}}{2021}]%
        {luo2021clip4clip}
\bibfield{author}{\bibinfo{person}{Huaishao Luo}, \bibinfo{person}{Lei Ji},
  \bibinfo{person}{Ming Zhong}, \bibinfo{person}{Yang Chen},
  \bibinfo{person}{Wen Lei}, \bibinfo{person}{Nan Duan}, {and}
  \bibinfo{person}{Tianrui Li}.} \bibinfo{year}{2021}\natexlab{}.
\newblock \showarticletitle{Clip4clip: An empirical study of clip for end to
  end video clip retrieval}.
\newblock \bibinfo{journal}{\emph{arXiv preprint arXiv:2104.08860}}
  (\bibinfo{year}{2021}).
\newblock


\bibitem[\protect\citeauthoryear{Mu, Kirillov, Wagner, and Xie}{Mu
  et~al\mbox{.}}{2021}]%
        {mu2021slip}
\bibfield{author}{\bibinfo{person}{Norman Mu}, \bibinfo{person}{Alexander
  Kirillov}, \bibinfo{person}{David Wagner}, {and} \bibinfo{person}{Saining
  Xie}.} \bibinfo{year}{2021}\natexlab{}.
\newblock \showarticletitle{SLIP: Self-supervision meets Language-Image
  Pre-training}.
\newblock \bibinfo{journal}{\emph{arXiv preprint arXiv:2112.12750}}
  (\bibinfo{year}{2021}).
\newblock


\bibitem[\protect\citeauthoryear{Nilsback and Zisserman}{Nilsback and
  Zisserman}{2008}]%
        {nilsback2008automated}
\bibfield{author}{\bibinfo{person}{Maria-Elena Nilsback} {and}
  \bibinfo{person}{Andrew Zisserman}.} \bibinfo{year}{2008}\natexlab{}.
\newblock \showarticletitle{Automated flower classification over a large number
  of classes}. In \bibinfo{booktitle}{\emph{2008 Sixth Indian Conference on
  Computer Vision, Graphics \& Image Processing}}. IEEE,
  \bibinfo{pages}{722--729}.
\newblock


\bibitem[\protect\citeauthoryear{Parkhi, Vedaldi, Zisserman, and
  Jawahar}{Parkhi et~al\mbox{.}}{2012}]%
        {parkhi2012cats}
\bibfield{author}{\bibinfo{person}{Omkar~M Parkhi}, \bibinfo{person}{Andrea
  Vedaldi}, \bibinfo{person}{Andrew Zisserman}, {and} \bibinfo{person}{CV
  Jawahar}.} \bibinfo{year}{2012}\natexlab{}.
\newblock \showarticletitle{Cats and dogs}. In \bibinfo{booktitle}{\emph{2012
  IEEE conference on computer vision and pattern recognition}}. IEEE,
  \bibinfo{pages}{3498--3505}.
\newblock


\bibitem[\protect\citeauthoryear{Radford, Kim, Hallacy, Ramesh, Goh, Agarwal,
  Sastry, Askell, Mishkin, Clark, et~al\mbox{.}}{Radford et~al\mbox{.}}{2021}]%
        {radford2021learning}
\bibfield{author}{\bibinfo{person}{Alec Radford}, \bibinfo{person}{Jong~Wook
  Kim}, \bibinfo{person}{Chris Hallacy}, \bibinfo{person}{Aditya Ramesh},
  \bibinfo{person}{Gabriel Goh}, \bibinfo{person}{Sandhini Agarwal},
  \bibinfo{person}{Girish Sastry}, \bibinfo{person}{Amanda Askell},
  \bibinfo{person}{Pamela Mishkin}, \bibinfo{person}{Jack Clark},
  {et~al\mbox{.}}} \bibinfo{year}{2021}\natexlab{}.
\newblock \showarticletitle{Learning transferable visual models from natural
  language supervision}. In \bibinfo{booktitle}{\emph{International Conference
  on Machine Learning}}. PMLR, \bibinfo{pages}{8748--8763}.
\newblock


\bibitem[\protect\citeauthoryear{Rao, Zhao, Chen, Tang, Zhu, Huang, Zhou, and
  Lu}{Rao et~al\mbox{.}}{2021}]%
        {rao2021denseclip}
\bibfield{author}{\bibinfo{person}{Yongming Rao}, \bibinfo{person}{Wenliang
  Zhao}, \bibinfo{person}{Guangyi Chen}, \bibinfo{person}{Yansong Tang},
  \bibinfo{person}{Zheng Zhu}, \bibinfo{person}{Guan Huang},
  \bibinfo{person}{Jie Zhou}, {and} \bibinfo{person}{Jiwen Lu}.}
  \bibinfo{year}{2021}\natexlab{}.
\newblock \showarticletitle{DenseCLIP: Language-Guided Dense Prediction with
  Context-Aware Prompting}.
\newblock \bibinfo{journal}{\emph{arXiv preprint arXiv:2112.01518}}
  (\bibinfo{year}{2021}).
\newblock


\bibitem[\protect\citeauthoryear{Ren, He, Girshick, and Sun}{Ren
  et~al\mbox{.}}{2015}]%
        {ren2015faster}
\bibfield{author}{\bibinfo{person}{Shaoqing Ren}, \bibinfo{person}{Kaiming He},
  \bibinfo{person}{Ross Girshick}, {and} \bibinfo{person}{Jian Sun}.}
  \bibinfo{year}{2015}\natexlab{}.
\newblock \showarticletitle{Faster r-cnn: Towards real-time object detection
  with region proposal networks}.
\newblock \bibinfo{journal}{\emph{Advances in neural information processing
  systems}}  \bibinfo{volume}{28} (\bibinfo{year}{2015}).
\newblock


\bibitem[\protect\citeauthoryear{Ridnik, Sharir, Ben-Cohen, Ben-Baruch, and
  Noy}{Ridnik et~al\mbox{.}}{2021}]%
        {ridnik2021ml}
\bibfield{author}{\bibinfo{person}{Tal Ridnik}, \bibinfo{person}{Gilad Sharir},
  \bibinfo{person}{Avi Ben-Cohen}, \bibinfo{person}{Emanuel Ben-Baruch}, {and}
  \bibinfo{person}{Asaf Noy}.} \bibinfo{year}{2021}\natexlab{}.
\newblock \showarticletitle{ML-Decoder: Scalable and Versatile Classification
  Head}.
\newblock \bibinfo{journal}{\emph{arXiv preprint arXiv:2111.12933}}
  (\bibinfo{year}{2021}).
\newblock


\bibitem[\protect\citeauthoryear{Sharma, Ding, Goodman, and Soricut}{Sharma
  et~al\mbox{.}}{2018}]%
        {sharma2018conceptual}
\bibfield{author}{\bibinfo{person}{Piyush Sharma}, \bibinfo{person}{Nan Ding},
  \bibinfo{person}{Sebastian Goodman}, {and} \bibinfo{person}{Radu Soricut}.}
  \bibinfo{year}{2018}\natexlab{}.
\newblock \showarticletitle{Conceptual captions: A cleaned, hypernymed, image
  alt-text dataset for automatic image captioning}. In
  \bibinfo{booktitle}{\emph{Proceedings of the 56th Annual Meeting of the
  Association for Computational Linguistics (Volume 1: Long Papers)}}.
  \bibinfo{pages}{2556--2565}.
\newblock


\bibitem[\protect\citeauthoryear{Shen, Li, Tan, Bansal, Rohrbach, Chang, Yao,
  and Keutzer}{Shen et~al\mbox{.}}{2021}]%
        {shen2021much}
\bibfield{author}{\bibinfo{person}{Sheng Shen}, \bibinfo{person}{Liunian~Harold
  Li}, \bibinfo{person}{Hao Tan}, \bibinfo{person}{Mohit Bansal},
  \bibinfo{person}{Anna Rohrbach}, \bibinfo{person}{Kai-Wei Chang},
  \bibinfo{person}{Zhewei Yao}, {and} \bibinfo{person}{Kurt Keutzer}.}
  \bibinfo{year}{2021}\natexlab{}.
\newblock \showarticletitle{How Much Can CLIP Benefit Vision-and-Language
  Tasks?}
\newblock \bibinfo{journal}{\emph{arXiv preprint arXiv:2107.06383}}
  (\bibinfo{year}{2021}).
\newblock


\bibitem[\protect\citeauthoryear{Tan and Bansal}{Tan and Bansal}{2019}]%
        {tan2019lxmert}
\bibfield{author}{\bibinfo{person}{Hao Tan} {and} \bibinfo{person}{Mohit
  Bansal}.} \bibinfo{year}{2019}\natexlab{}.
\newblock \showarticletitle{Lxmert: Learning cross-modality encoder
  representations from transformers}.
\newblock \bibinfo{journal}{\emph{arXiv preprint arXiv:1908.07490}}
  (\bibinfo{year}{2019}).
\newblock


\bibitem[\protect\citeauthoryear{Tejankar, Wu, Xie, Khabsa, Pirsiavash, and
  Firooz}{Tejankar et~al\mbox{.}}{2021}]%
        {tejankar2021fistful}
\bibfield{author}{\bibinfo{person}{Ajinkya Tejankar}, \bibinfo{person}{Bichen
  Wu}, \bibinfo{person}{Saining Xie}, \bibinfo{person}{Madian Khabsa},
  \bibinfo{person}{Hamed Pirsiavash}, {and} \bibinfo{person}{Hamed Firooz}.}
  \bibinfo{year}{2021}\natexlab{}.
\newblock \showarticletitle{A Fistful of Words: Learning Transferable Visual
  Models from Bag-of-Words Supervision}.
\newblock \bibinfo{journal}{\emph{arXiv preprint arXiv:2112.13884}}
  (\bibinfo{year}{2021}).
\newblock


\bibitem[\protect\citeauthoryear{Van~den Oord, Li, Vinyals,
  et~al\mbox{.}}{Van~den Oord et~al\mbox{.}}{2018}]%
        {van2018representation}
\bibfield{author}{\bibinfo{person}{Aaron Van~den Oord}, \bibinfo{person}{Yazhe
  Li}, \bibinfo{person}{Oriol Vinyals}, {et~al\mbox{.}}}
  \bibinfo{year}{2018}\natexlab{}.
\newblock \showarticletitle{Representation learning with contrastive predictive
  coding}.
\newblock \bibinfo{journal}{\emph{arXiv preprint arXiv:1807.03748}}
  \bibinfo{volume}{2}, \bibinfo{number}{3} (\bibinfo{year}{2018}),
  \bibinfo{pages}{4}.
\newblock


\bibitem[\protect\citeauthoryear{Vaswani, Shazeer, Parmar, Uszkoreit, Jones,
  Gomez, Kaiser, and Polosukhin}{Vaswani et~al\mbox{.}}{2017}]%
        {vaswani2017attention}
\bibfield{author}{\bibinfo{person}{Ashish Vaswani}, \bibinfo{person}{Noam
  Shazeer}, \bibinfo{person}{Niki Parmar}, \bibinfo{person}{Jakob Uszkoreit},
  \bibinfo{person}{Llion Jones}, \bibinfo{person}{Aidan~N Gomez},
  \bibinfo{person}{{\L}ukasz Kaiser}, {and} \bibinfo{person}{Illia
  Polosukhin}.} \bibinfo{year}{2017}\natexlab{}.
\newblock \showarticletitle{Attention is all you need}.
\newblock \bibinfo{journal}{\emph{Advances in neural information processing
  systems}}  \bibinfo{volume}{30} (\bibinfo{year}{2017}).
\newblock


\bibitem[\protect\citeauthoryear{Xiao, Hays, Ehinger, Oliva, and Torralba}{Xiao
  et~al\mbox{.}}{2010}]%
        {xiao2010sun}
\bibfield{author}{\bibinfo{person}{Jianxiong Xiao}, \bibinfo{person}{James
  Hays}, \bibinfo{person}{Krista~A Ehinger}, \bibinfo{person}{Aude Oliva},
  {and} \bibinfo{person}{Antonio Torralba}.} \bibinfo{year}{2010}\natexlab{}.
\newblock \showarticletitle{Sun database: Large-scale scene recognition from
  abbey to zoo}. In \bibinfo{booktitle}{\emph{2010 IEEE computer society
  conference on computer vision and pattern recognition}}. IEEE,
  \bibinfo{pages}{3485--3492}.
\newblock


\bibitem[\protect\citeauthoryear{Zhang, Li, Hu, Yang, Zhang, Wang, Choi, and
  Gao}{Zhang et~al\mbox{.}}{2021b}]%
        {zhang2021vinvl}
\bibfield{author}{\bibinfo{person}{Pengchuan Zhang}, \bibinfo{person}{Xiujun
  Li}, \bibinfo{person}{Xiaowei Hu}, \bibinfo{person}{Jianwei Yang},
  \bibinfo{person}{Lei Zhang}, \bibinfo{person}{Lijuan Wang},
  \bibinfo{person}{Yejin Choi}, {and} \bibinfo{person}{Jianfeng Gao}.}
  \bibinfo{year}{2021}\natexlab{b}.
\newblock \showarticletitle{Vinvl: Revisiting visual representations in
  vision-language models}. In \bibinfo{booktitle}{\emph{Proceedings of the
  IEEE/CVF Conference on Computer Vision and Pattern Recognition}}.
  \bibinfo{pages}{5579--5588}.
\newblock


\bibitem[\protect\citeauthoryear{Zhang, Cheng, Huang, Wen, Feng, Li, and
  Guo}{Zhang et~al\mbox{.}}{2021a}]%
        {zhang2021simple}
\bibfield{author}{\bibinfo{person}{Youcai Zhang}, \bibinfo{person}{Yuhao
  Cheng}, \bibinfo{person}{Xinyu Huang}, \bibinfo{person}{Fei Wen},
  \bibinfo{person}{Rui Feng}, \bibinfo{person}{Yaqian Li}, {and}
  \bibinfo{person}{Yandong Guo}.} \bibinfo{year}{2021}\natexlab{a}.
\newblock \showarticletitle{Simple and Robust Loss Design for Multi-Label
  Learning with Missing Labels}.
\newblock \bibinfo{journal}{\emph{arXiv preprint arXiv:2112.07368}}
  (\bibinfo{year}{2021}).
\newblock


\end{thebibliography}

\clearpage
\newpage
\appendix
\section{Appendix}

\subsection{IDEA Algorithm}
\label{sec:pseudo}

We present the pseudo-code of IDEA training algorithm in Algorithm~\ref{alg:pseudo}. IDEA jointly trains Image-Text Contrastive learning (ITC) with the text and Multi-Label Recognition (MLR) with the tags extracted from the text according to the tag list. The visual global embedding corresponds to the text and the visual spatial embeddings correspond to the tags. In addition, IDEA can identify the additional missing tags in an online manner by leveraging the simple missing tag recognition loss.

\vspace{-0.5em}
\begin{algorithm}[h]
\textcolor{xinyu}{\# fi,\ ft:\ image, text encoders }  \\ 
\textcolor{xinyu}{\# hi,\ ht:\ image, text projectors} \\
\textcolor{xinyu}{\# tag:\  image tags extracted from text,~(binary encoding)} \\
\textcolor{xinyu}{\# epoch:\ current training epoch} \\
\textcolor{pink}{def}\ forward\,(\,img,\ text,\ tag,\ epoch\,): \\
\qquad   \textcolor{xinyu}{\#\ get image sptial feature xs and global feature xg}\\
\qquad   xs,\ xg = fi\,(\,img\,)   \\
\qquad   \textcolor{xinyu}{\#\ multi-label recognition}\\
\qquad   l\_mlr,\ tag,\,pseudo =\ mlr\,(\,xs,\ tag,\ epoch)\\
\qquad   \textcolor{xinyu}{\#\ convert tag to text}\\
\qquad   tag2text\ =\ \textcolor{pink}{Merge}\,(\,tag\,)  \\

\qquad   text = \textcolor{pink}{Cat}\,(\,text,\ tag2text\,) \\
\qquad   \textcolor{xinyu}{\#\ get text global feature yg}\\
\qquad   yg = ft\,(\,text\,)  \\
\qquad   \textcolor{xinyu}{\#\ image text contrastive learning}\\
\qquad   l\_itc = itc\,(\,xg,\ yg,\,pseudo\,) \\
\qquad   loss\ =\ l\_mlr\ +\ l\_itc \\
\qquad   \textcolor{pink}{return} loss 
\vspace{0.4em}\\
\textcolor{xinyu}{\# ml,\ ce:\ ML-Decoder,\, changing epoch}             \\
\textcolor{pink}{def}\ mlr\,(\,xs,\ tag,\ epoch\,): \\
\qquad   logit = ml\,(xs) \\  
\qquad   target = tag \\
\vspace{0.3em}
\qquad   if\ epoch\ >= ce:  \\
\qquad \qquad  target = \textcolor{pink}{Where}\,(\,Sigmoid\,(\,logit\,)\ >\ threshold\ ) \\

\vspace{0.3em}
\qquad   l\_mlr = \textcolor{pink}{BinaryCrossEntropy}\,(\,logit,\ target)\  \\
\qquad   pseudo = \textcolor{pink}{Where}\,(\,(target-tag).sum\,(dim=1)\,>0)

\qquad  \textcolor{pink}{return} l\_mlr,\ tag * pseudo,\ pseudo
\vspace{0.3em}
\textcolor{xinyu}{\# s:\ learnable temperature\\}   
\textcolor{pink}{def}\ itc\,(\,xg,\ yg,\,pseudo): \\
\qquad    zi = \textcolor{pink}{Normalize}\,(hi\,(\,xg\,))  \\
\qquad    zt = \textcolor{pink}{Normalize}\,(ht\,(\,yg\,))  
\vspace{0.3em}\\
\qquad    similarity = exp(s)\ *\ zi\ @\ zt.T \\
\qquad    target = \textcolor{pink}{Cat}\,(\,range\,(\,N\,),\ pseudo,) \\
\vspace{0.3em}
\qquad    li = \textcolor{pink}{KL}\,(\,similarity,\ target\,) \\
\qquad    lt = \textcolor{pink}{KL}\,(\,similarity.T,\ target.T\,) \\
\qquad    l\_itc = (\,li + lt\,) / 2 \\
\qquad    \textcolor{pink}{return} l\_itc

\caption{IDEA: PyTorch-like Pseudocode} 
\label{alg:pseudo}
\algorithmfootnote{\textbf{Notes}: @:matrix multiplication operator; cat:concatenation. }
\end{algorithm}

\qquad

\qquad












\vspace{-2.0em}
\subsection{Prompt Engineering and Ensembling.}
\begin{figure}[ht]
\vspace{-1.0em}
  \centering
  \includegraphics[width=.45\textwidth]{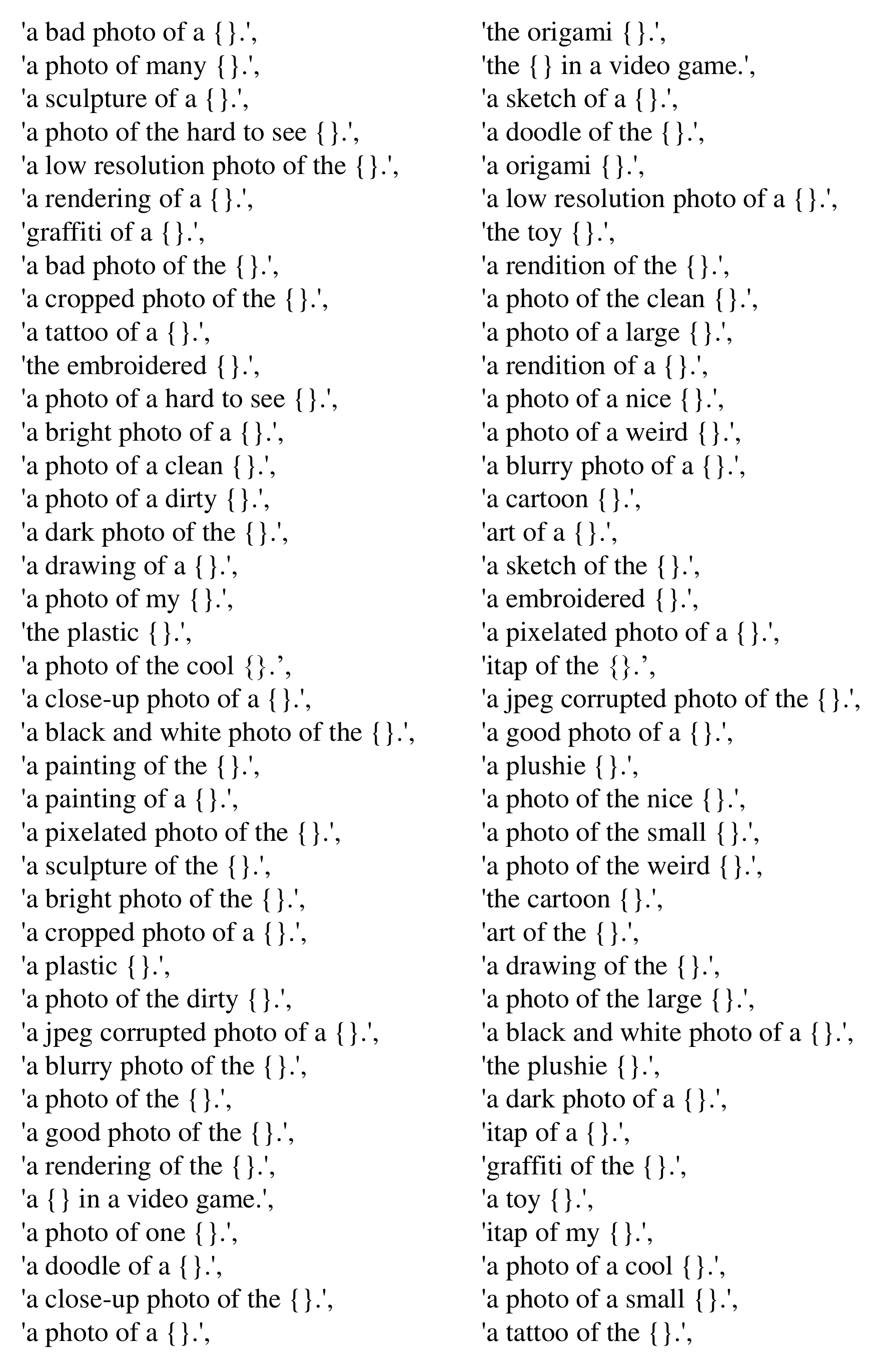} 
  \vspace{-1.0em}
  \caption{Prompt templates for zero-shot evaluation.}
  \vspace{-1.0em}
  \label{fig:prompt_templete}
\end{figure}

Following~\cite{radford2021learning}, we adopt prompt engineering and ensembling to improve the zero-shot evaluation on downstream datasets except for DTD~\cite{cimpoi2014describing} and OxfordPets~\cite{parkhi2012cats}. The prompt templates are shown in Figure~\ref{fig:prompt_templete}. We only adopt the prompt template {\rm{"a photo of a \{label\}"}} for DTD and OxfordPets, since we found prompt ensembling with the prompt templates in Figure~\ref{fig:prompt_templete} will harm the performance.

\newpage
\subsection{More Example Visualization Results}
\label{sec:more_visualization}
\begin{figure*} [hb]
\centering
  \includegraphics[width=1\linewidth]{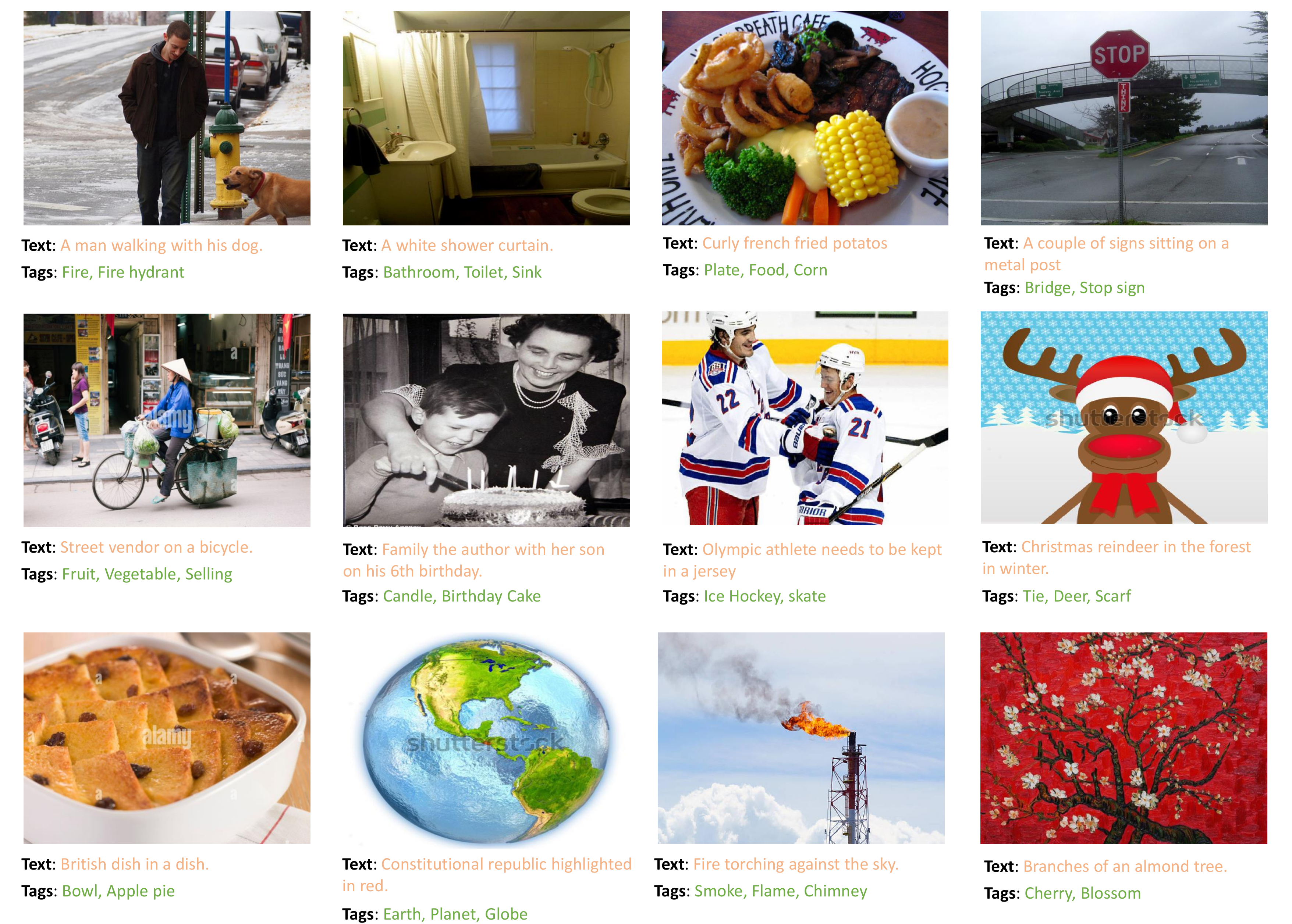}
  \caption{More example visualization results. Text refers to the original co-occurrent texts with the image. Tags refer to the identified tags by IDEA, including objects, scenes, attributes, actions, etc. These image tags are entirely learned from the texts and recognized online.}
  \vspace{-0.5em}
  \label{fig:visualization_app}
\end{figure*}

\end{document}